\PassOptionsToPackage{unicode}{hyperref}
\PassOptionsToPackage{hyphens}{url}
\documentclass[11pt]{article}
\usepackage{amsmath,amssymb}
\usepackage{iftex}
\ifPDFTeX
  \usepackage[T1]{fontenc}
  \usepackage[utf8]{inputenc}
  \usepackage{textcomp} % provide euro and other symbols
\else % if luatex or xetex
  \usepackage{unicode-math} % this also loads fontspec
  \defaultfontfeatures{Scale=MatchLowercase}
  \defaultfontfeatures[\rmfamily]{Ligatures=TeX,Scale=1}
\fi
\usepackage{lmodern}
\ifPDFTeX\else
\fi
\IfFileExists{upquote.sty}{\usepackage{upquote}}{}
\IfFileExists{microtype.sty}{% use microtype if available
  \usepackage[]{microtype}
  \UseMicrotypeSet[protrusion]{basicmath} % disable protrusion for tt fonts
}{}
\makeatletter
\@ifundefined{KOMAClassName}{% if non-KOMA class
  \IfFileExists{parskip.sty}{%
    \usepackage{parskip}
  }{% else
    \setlength{\parindent}{0pt}
    \setlength{\parskip}{6pt plus 2pt minus 1pt}}
}{% if KOMA class
  \KOMAoptions{parskip=half}}
\makeatother
\usepackage{xcolor}
\usepackage{longtable,booktabs,array}
\usepackage{calc} % for calculating minipage widths
\usepackage{etoolbox}
\makeatletter
\patchcmd\longtable{\par}{\if@noskipsec\mbox{}\fi\par}{}{}
\makeatother
\IfFileExists{footnotehyper.sty}{\usepackage{footnotehyper}}{\usepackage{footnote}}
\makesavenoteenv{longtable}
\usepackage{graphicx}
\usepackage{geometry}
\makeatletter
\def\maxwidth{\ifdim\Gin@nat@width>\linewidth\linewidth\else\Gin@nat@width\fi}
\def\maxheight{\ifdim\Gin@nat@height>\textheight\textheight\else\Gin@nat@height\fi}
\makeatother
\setkeys{Gin}{width=\maxwidth,height=\maxheight,keepaspectratio}
\makeatletter
\def\fps@figure{htbp}
\makeatother
\ifLuaTeX
  \usepackage{luacolor}
  \usepackage[soul]{lua-ul}
\else
  \usepackage{soul}
\fi
\ifLuaTeX
  \usepackage{selnolig}  % disable illegal ligatures
\fi
\usepackage{bookmark}
\IfFileExists{xurl.sty}{\usepackage{xurl}}{} % add URL line breaks if available
\hypersetup{
  pdftitle={Argument-Aware Semantic Alignment of Normative Texts: A Toulmin-Based Neuro-Symbolic Approach},
  pdfauthor={Dr. William Schroeder},
  hidelinks,
  pdfcreator={LaTeX via pandoc}}

\title{Argument-Aware Semantic Alignment of Normative Texts: A Toulmin-Based Neuro-Symbolic Approach}
\author{Dr. William Schroeder\\Purdue University; CTO Cleantech Software\\ORCID: 0009-0009-7703-9951\\\href{mailto:bschroeder@cleantechsoftware.com}{\nolinkurl{bschroeder@cleantechsoftware.com}}}
\date{August 21, 2026}

\usepackage{pdflscape}
\usepackage{float}
\begin{document}
\maketitle

\section*{Abstract}\label{abstract}

Semantic alignment between specialized normative texts is challenging
when equivalent requirements are expressed using different terminology,
syntactic constructions, and levels of abstraction. Existing approaches
to requirements and standards alignment commonly rely on lexical
overlap, distributional representation, or semantic similarity. While
effective at capturing topical and conceptual relatedness, these
representations can overlook the logical argumentative structure through
which normative claims are supported, qualified, and justified.

This paper investigates whether explicit argument structure provides
information complementary to neural semantic representations for
aligning normative requirements. We formulate cross-standard control
mapping as an argument-aware semantic alignment task and develop a
neuro-symbolic pipeline that combines neural representations of text
with structured features derived from the Toulmin argumentation
framework. An LLM-based argument explicitation procedure identifies
claims, grounds, warrants, qualifiers, and backing as well as
reconstructs enthymemes from normative requirements. These
representations are subsequently incorporated into a neuro-symbolic
alignment model through argument aware similarity and structural
features.

We evaluate the approach on a benchmark of mappings between NERC-CIP and
NIST-CSF controls. The results show that incorporating argument-derived
representations improves alignment performance over a neuro-symbolic
semantic baseline. Feature selection experiments indicate that warrant
related representations provide particularly strong predictive signal,
suggesting that the relationship between a normative claim and the
reasoning supporting that claim contains information not captured by
conventional semantic similarity alone. A compact representation
centered on claims, grounds, and warrants also performs competitively
with the larger argument-derived feature set, indicating that the
benefits of argument structure do not necessarily require a complete
Toulmin representation.

These findings provide preliminary evidence that argumentative structure
can function as a useful intermediate representation for semantic
alignment of specialized normative texts. The study uses cybersecurity
standards as a controlled testbed rather than claiming
domain-independent generalization. These results should be interpreted
as evidence of the downstream utility of argument-derived intermediate
representations and the resulting argument graphs generated through LLM
based explicitation may also support future work on retrieval,
reasoning, and explanation generation over normative text.

\section{Introduction}\label{introduction}

Cybersecurity standards and frameworks form a primary defense for
critical infrastructure (Schneier, 2018). In the United States, energy
systems are designated Critical Infrastructure and Key Resources under
HSPD-7 (U.S. Department of Homeland Security, 2010). Because no single
standard fully covers complex critical infrastructure environments,
multiple overlapping frameworks must be synthesized. This harmonization
process is essential to reduce the attack surface within complex
critical infrastructure environments and also provide an industry
relevant testbed for semantic alignment between specialized texts.

Standards are normative documents which contain different terminology,
organization structure, and levels of abstraction. Determining whether
two requirements express substantially equivalent controls is a problem
of semantic alignment between specialized texts. Prior work framed
harmonization as a semantic analysis task and applied three
architectural approaches: symbolic approaches that encode standards into
logical graph structures grounded by taxonomies or ontologies (Olifer et
al., 2019); sub-symbolic approaches that rely on vector embeddings or
large language models (Agarwal et al., 2022); and hybrid neuro-symbolic
(NeSy) architectures that combine the pattern matching strength of
neural networks and the interpretability of symbolic graphs. When
integrated into harmonization classifiers these approaches achieve F1
scores approaching human-expert agreement (Abdeen et al., 2023;
Schroeder, 2025).

However, two fundamental gaps persist. First, existing models capture
surface semantic and symbolic similarity but overlook inferential
relationships that structure regulatory text. Second, the resulting
models provide limited access to a human-interpretable justification,
constraining adoption by the diverse technical, policy, and liability
stakeholders who must jointly validate compliance decisions. This
research addresses both gaps by reframing standards as structured
arguments and operationalizing the Toulmin argumentation model, see
Figure 1. Toulmin argument roles supply discriminative features that
enrich classification while producing structured argument graphs that
lay the architectural foundation for future explanation generation. The
Toulmin schema is integrated into an open-source neuro-symbolic pipeline
that augments prediction with argumentation derived signal (Lawrence,
J., \& Reed, C., 2019).

This research is based on the theory that in domains which are organized
around explicit or implicit reasoning, leveraging argumentation mining
(AM) techniques yields richer semantic representations. From this
theoretical foundation we hypothesize that Toulmin argument roles carry
complementary predictive signal to conventional semantic representations
and constitute a natural foundation for human-interpretable
harmonization. To investigate this hypothesis, an NLP argumentation
mining pipeline was developed to automate cybersecurity standard AM
annotation, rank the AM feature importance, and measure the pipeline's
classification accuracy. Three primary research questions directed the
research.

\begin{quote}
RQ1: Does Argument Structure Improve Harmonization Performance?
\emph{What is the effect on cross-validation and test F1 scores of
integrating Toulmin-derived argument mining features into a
neuro-symbolic NLP pipeline for NERC-CIP and NIST-CSF standards
harmonization, compared to the baseline pipeline without these
features?}

RQ2: Does Annotation Methodology Matter? \emph{Does LLM-based Toulmin
enthymeme explicitation produce statistically significant higher F1
scores than traditional lexical pattern-based extraction when
identifying and explicating argumentative structures in cybersecurity
standards documents?}

RQ3: How Much Argumentation Structure Is Required? \emph{What are the
empirical feature importance rankings of Toulmin argument components and
to what extent do they exhibit stable, reproducible patterns?}
\end{quote}

The contributions this research makes are (1) empirical validation of a
fused NeSy+AM-cosine model that yields statistically significant CV-F1
and Test-F1 improvements, replicated across independent dataset splits,
(2) empirical comparison of annotation methodology which isolated and
verified the statistical significance of enthymeme reconstruction, (3) a
reusable NLP AM pipeline and testing framework released as open source
to support future research.

\section{Background}\label{background}

\subsection{Prior Work}\label{prior-work}

Standards harmonization is an important and well documented applied
problem in the cybersecurity domain. Eggers et al. (2018) highlighted
the potential of applying computational linguistic techniques to
cybersecurity standards automation to reduce costs, improve
reproducibility, and increase transparency. Since cybersecurity
standards and frameworks form the primary defense for critical
infrastructure, techniques that improve the harmonization automation
process are essential tools in reducing the attack surface within
complex critical infrastructure environments (Schneier, 2018).

Normative requirements such as cybersecurity standards, frequently
contain implicit argumentative structures. A requirement may state a
claim about what an organization must do, provide grounds describing the
relevant system or risk, specify conditions or qualifications, and
implicitly or explicitly encode a warrant connecting those grounds to
the required action. Although these structures are often not expressed
as explicit arguments they can be represented this way using established
argumentation frameworks such as the Toulmin model. The field of
argument mining provides computational methods for identifying Toulmin
argumentation structures in natural language but primarily treats
argumentative structure as the primary prediction target (Lawrence, J.,
\& Reed, C., 2019). In contrast, we investigate a complementary use of
argument structure as an intermediate semantic structure that can be
used in the normative text alignment task of standards harmonization.

Semantic alignment involves determining whether two textual units
express sufficiently related meaning to support a downstream task such
as retrieval, classification, or knowledge integration. Conventional
approaches range from lexical similarity and information retrieval
techniques to distributed representations produced by neural language
models. Sentence and document embedding models have substantially
improved the ability to identify semantically related text despite
lexical variation. Cross-encoders and other neural architectures can
additionally model interactions between paired texts. However, normative
text introduces additional challenges of varying levels of explicitness,
granularity, scope and justification. Our work specifically investigates
standards requirements as a specialized form of normative text and
whether the structured argumentative representations within them can
provide complementary information in this environment.

Subsequent work in this area has explored three main architectural
approaches to harmonization automation. Specifically, Adam et al. (2019)
used neural approaches to classify security requirements as corporate
control labels, Agarwal et al. (2022) and Bar-Haim et al. (2023)
extended this approach by fine-tuning a RoBERTa model and adding an
intermediate taxonomy layer. Olifer et al (2019) used a symbolic
approach to represent security standards as knowledge graphs and measure
similarity through graph distances and isomorphisms. Gogireddy et al,
(2024) proposed a generalized hybrid neuro-symbolic system capable of
leveraging each of their complementary strengths. These approaches
captured semantic and symbolic similarity but overlooked the inferential
relationships that inherently structure the regulatory domain. This
research hypothesizes that these relationships can be modeled as formal
Toulmin arguments which yield additional discriminative signal. This
research leverages cybersecurity standards harmonization as an
important, applied problem that provides a controlled testbed for
applying argument mining to normative text rather than claiming
domain-independent generalization

\subsection{Toulmin Argumentation
Mining}\label{toulmin-argumentation-mining}

Argument mining is concerned with automatically identifying
argumentative structures within natural language text. Several
theoretical models have been operationalized as frameworks (e.g.,
ASPIC+, Dung, Toulmin, Walton), that model argumentation components
(e.g., claims, grounds, warrant, evidence, qualifier, etc.) and their
relationships, see Figure 1 below. Empirical precedent exists in
adjacent domains for using argumentation structure to enhance
classification and explainability. For example, Haley et al. (2008)
applied Toulmin structures to validate security requirements and Viger
et al. (2023) adopted the Toulmin structure as a mapping ontology to
Assurance Cases (ACs) and Goal Structured Notation (GSN) used by ISO
21434 within the safety domain. The Toulmin framework was adopted for
our research for its balance of structure and flexibility, its
widespread use in computational linguistics, and empirical precedent in
adjacent domains (Lytos et al., 2019).

\begin{figure}[H]
\centering
\includegraphics[width=0.85\linewidth,height=0.5\textheight,keepaspectratio]{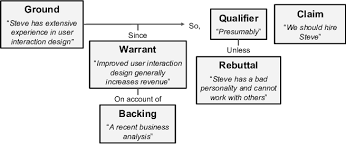}
\caption{Toulmin Argument Taxonomy}
\label{fig:toulmin-taxonomy}
\par\smallskip
\noindent\begin{minipage}{0.92\linewidth}\small\textit{Note:} Anatomy of Community Opinions on Usability Issues Using Argumentation Models, by W. Wang et al., 2020, ResearchGate. \href{https://www.researchgate.net/figure/Relationship-among-the-six-components-of-the-Toulmins-model-of-argumentation-59-with_fig1_338688363}{ResearchGate figure page}. Copyright 2020 by the authors.\end{minipage}
\end{figure}

An important component and recurring challenge within argument mining is
that the relationships between components are frequently implicit.
Normative language often states a required action and its relevant
context without explicitly stating the reasoning connecting them. For
example, where a requirement describes a class of systems and specifies
a security control, the warrant may represent the relationships between
the relevant security risk and the required control.

To address this challenge, recent work has explored methods for
reconstructing enthymemes explicitly. One technique proposed by Gupta et
al. (2024) is LLM-explicitation. Gupta demonstrated that LLMs could
effectively decompose informal arguments into Toulmin
claim-reason-warrant triples. As part of our research, we leverage a
similar LLM-explicitation annotation approach to cybersecurity standards
and empirically tests the approach to determine its effectiveness. We
then use the most effective empirically validated annotation approach
within our hypothesis test that argumentation features complement
Neuro-symbolic harmonization classification.

\subsection{Task Definition}\label{task-definition}

We formulate normative control alignment as a binary semantic alignment problem. Given two normative requirements, $r_i$ and $r_j$, the task is to determine whether they constitute a valid alignment according to the OLIR mapping annotation. The conventional representation of this problem is based primarily on the semantic representations of $r_i$ and $r_j$:

\begin{equation}
s(r_i,r_j) = (E(r_i),E(r_j)),
\end{equation}
where $E(\cdot)$ denotes a neural text embedding and $s(\cdot)$ is a similarity function. Our research hypothesizes that Toulmin argument roles carry complementary predictive signal to conventional semantic representations. We therefore introduce an argument representation,

\begin{equation}
A(r) = \{a_1,a_2,\ldots,a_n\},
\end{equation}
which represents $n$ Toulmin argumentation features. Alignment is then modeled using both semantic and argument-derived information,

\begin{equation}
S(r_i,r_j) = f(x_1,x_2,\ldots,x_n).
\end{equation}
The hypothesis, therefore, is not that argumentative structure replaces semantic representations, but that it provides complementary information that can improve alignment.
\section{Methods}\label{methods}

\subsection{Dataset and Annotation}\label{dataset-and-annotation}

All datasets and schemas adhere to open science protocols to enable
replication and promote reuse of the core research artifacts. The D1 and
D2 datasets described in this section are publicly available or
reproducible from publicly available information. Further, the
argumentation annotation layer, prompt schema, dataset construction
code, and annotations are also publicly available upon request, see
Appendix A.

\paragraph{OLIR NERC-CIP/NIST CSF Dataset
(D1)}\label{olir-nerc-cipnist-csf-dataset-d1}

This research uses the open source NIST OLIR dataset for all
experiments. OLIR provides expert annotated harmonized pairs of NERC-CIP
5.1 requirements and NIST-CSF 1.1 subcategories. The OLIR raw dataset is
published in document 90 from NIST and contains 324 explicitly labeled
harmonized pairs (National Institute of Standards and Technology,
2023)..

Each harmonized pair is also labeled with a Strength of Relationship
(SoR) score on an ordinal scale ranging from partial support (SoR=2) to
an exact or near match (SoR=8) Only moderate to strong OLIR
(SoR\textgreater2) annotations are labeled as harmonizable. Annotations
with limited or weak overlap (SoR\textless=2) are excluded as positive
class labels. This threshold reflects a deliberate task definition
choice to create an audit-quality, defeasible, compliance mapping.
Cleaning the data and applying these criteria transforms the OLIR raw
dataset (dataset\_raw\_x.csv) into an intermediate dataset
(dataset\_xformed\_x.csv).

OLIR explicitly assesses only standard pairs with a strength of
relationship degree (SoR), making the dataset label-asymmetric.
Therefore, it is necessary to create synthetic negative samples to
balance the labeled sets. A hybrid sampling strategy using both hard
semantic and stochastic rule based sampling techniques (70\% hard
semantic, 30\% rule-based stochastic) was used to guarantee both
balanced label coverage of the generated samples and ensure semantic
validity.

Two composable factory classes were implemented to generate the required
synthetic data. A Hard Semantic Negative Factory class constructs
synthetic pairs by selecting up to the top-k limit (5) of highest
embedding cosine similarity standards not mapped as harmonizable. This
design assures semantically hard sample pairs are generated that share
vocabulary and domain categories but are not assessed as harmonizable by
OLIR. Training the classifier with semantically hard samples improves
the generalizability of the model and prevents the model from learning
non-generalizable trivial lexical patterns. The Negative Sample Rule
Factory randomly pairs NIST CSF and NERC CIP requirements, excluding any
OLIR harmonized pairs labeled with a SoR. Rule-based pairs represent
negative synthetic samples with higher semantic contrast than those
generated by the Hard Semantic Negative Factory.

All pairs were deduplicated within and across factory types, both types
were combined in a hybrid 70/30 ratio and shuffled with a fixed random
seed before concatenating with the positive samples. Generating
synthetic negative samples and then balancing the dataset 1:1 between
positive and negative pairs yields an audit-quality, defeasible,
compliance mapping contained in the golden D1 input dataset
(dataset\_balanced\_x.csv), see Table 1 below.

\begin{longtable}[]{@{}
  >{\raggedright\arraybackslash}p{(\columnwidth - 6\tabcolsep) * \real{0.2404}}
  >{\raggedright\arraybackslash}p{(\columnwidth - 6\tabcolsep) * \real{0.1971}}
  >{\raggedright\arraybackslash}p{(\columnwidth - 6\tabcolsep) * \real{0.1858}}
  >{\raggedright\arraybackslash}p{(\columnwidth - 6\tabcolsep) * \real{0.3767}}@{}}
\caption{Dataset D1}\tabularnewline
\toprule\noalign{}
\begin{minipage}[b]{\linewidth}\raggedright
Split
\end{minipage} & \begin{minipage}[b]{\linewidth}\raggedright
Positive
\end{minipage} & \begin{minipage}[b]{\linewidth}\raggedright
Synthetic negative
\end{minipage} & \begin{minipage}[b]{\linewidth}\raggedright
Total
\end{minipage} \\
\midrule\noalign{}
\endfirsthead
\toprule\noalign{}
\begin{minipage}[b]{\linewidth}\raggedright
Split
\end{minipage} & \begin{minipage}[b]{\linewidth}\raggedright
Positive
\end{minipage} & \begin{minipage}[b]{\linewidth}\raggedright
Synthetic negative
\end{minipage} & \begin{minipage}[b]{\linewidth}\raggedright
Total
\end{minipage} \\
\midrule\noalign{}
\endhead
\bottomrule\noalign{}
\endlastfoot
OLIR DS & 198 & 0 & 198 \\
Hard Semantic & 0 & 138 & 138 \\
Stochastic Rule Based & 0 & 60 & 60 \\
Total & 198 & 198 & 396 \\
\end{longtable}

\paragraph{Argument Annotated Dataset
(D2)}\label{argument-annotated-dataset-d2}

The golden OLIR dataset (D1) is transformed into a parallel
argumentation annotated dataset (D2). The annotation schema was based on
Toulmin argumentation theory and annotates each regulatory requirement
as a structured argument consisting of an extractable Claim, Grounds,
Warrant, Qualifier, and Backing component. Each role's regulatory
interpretation is defined in Table 2 below.

A framework was developed to automate the annotation by extending an
open source neuro-symbolic classifier with a symbolic argumentation
annotator class. The result is a complete annotated argument structure
for all OLIR standards cached in a persistent JSON store and hashed to
the original standards text for downstream retrieval and
reproducibility. Together, the balanced OLIR derived pairs (D1) and the
Toulmin annotated argument structures (D2) constitute the experimental
input data

\begin{longtable}[]{@{}
  >{\raggedright\arraybackslash}p{(\columnwidth - 2\tabcolsep) * \real{0.2259}}
  >{\raggedright\arraybackslash}p{(\columnwidth - 2\tabcolsep) * \real{0.7741}}@{}}
\caption{Annotation Role Definitions (D2)}\tabularnewline
\toprule\noalign{}
\begin{minipage}[b]{\linewidth}\raggedright
Role
\end{minipage} & \begin{minipage}[b]{\linewidth}\raggedright
Cybersecurity Standard Schema
\end{minipage} \\
\midrule\noalign{}
\endfirsthead
\toprule\noalign{}
\begin{minipage}[b]{\linewidth}\raggedright
Role
\end{minipage} & \begin{minipage}[b]{\linewidth}\raggedright
Cybersecurity Standard Schema
\end{minipage} \\
\midrule\noalign{}
\endhead
\bottomrule\noalign{}
\endlastfoot
Claim & The security objective or control goal the requirement
asserts \\
Grounds & The factual or evidentiary basis cited or implied \\
Warrant & The risk reasoning connecting grounds to the claim \\
Qualifier & Applicable scope, conditions, system boundaries \\
Backing & Regulatory authority, cited standards, references \\
\end{longtable}

Each standard is annotated by the AMAnnotator class which parses each
OLIR pair into its Toulmin components. However, since regulatory
requirements are not written as formal arguments this can result it
incomplete Toulmin structures. For example, backing or grounds may be
implicitly assumed as domain knowledge and not formally or explicitly
stated in the requirement. This is the well-known and actively
researched enthymeme issue in argument mining (AM). AMAnnotator
implements two common annotation methodologies that handle enthymemes
differently, extraction and explicitation

Extraction is done using the spacy NLP toolkit to directly parse the
requirement into Toulmin argument structures. Keyword and regex matching
are done to directly extract the exact text in the requirement as a
specific Toulmin component. If enthymemes are detected during the
text-extraction methodology they are annotated as such but are not made
explicit. In contrast, the text-explicitation methodology infers and
makes explicit all enthymemes. Text-explicitation uses one-shot
prompting via an API with llama-3.3-70b-versatile hosted on the Groq
platform with temperature=0 for deterministic results to infer and make
explicit all detected enthymemes, see Appendix A.

\subsection{\texorpdfstring{Experimental Setup and Evaluation
Protocol
}{Experimental Setup and Evaluation Protocol }}\label{experimental-setup-and-evaluation-protocol}

\paragraph{System Architecture}\label{system-architecture}

This research leverages and extends the capabilities of the Rosetta
neuro-symbolic (NeSy) NLP platform. This pre-existing platform natively
has the core capabilities required to address RQ1-RQ3, see Figure 2
below. This research leveraged and extended the Rosetta platform to
rapidly prototype the experiment protocol. Rosetta provides an open,
extensible NLP classification framework that encapsulates five pipeline
stages and their orchestration into six architectural framework
components. The core framework components: data processing, model
generation, feature extraction, classification, statistical analysis,
and pipeline orchestration were modified and extended to work with
argumentation structures. The combined modifications provide a research
contribution and a logical implementation of a Toulmin argumentation
extension to the platform (W. N. Schroeder, 2025).

\begin{figure}[H]  
\centering
\includegraphics[width=0.95\linewidth,height=0.82\textheight,keepaspectratio]{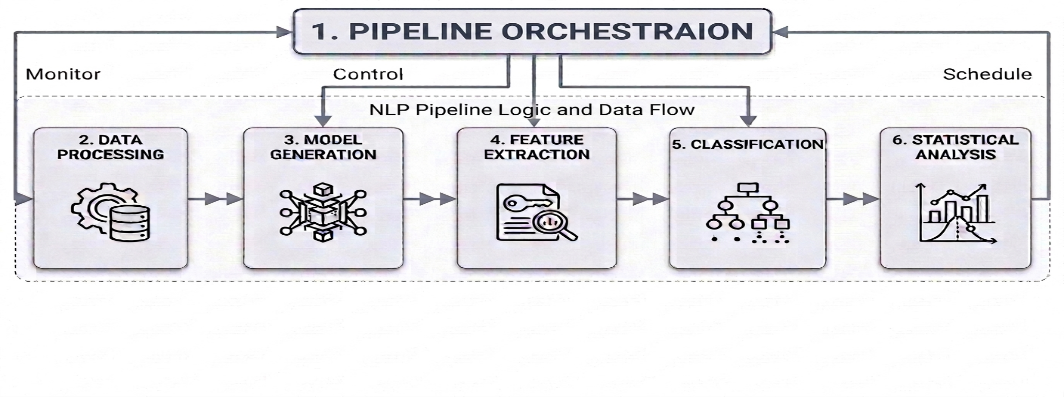}
\caption{Core Pipeline Logical and Data Flow}
\label{fig:core-pipeline}
\end{figure}

The following extensions implemented a symbolic argumentation model
based on the Toulmin argumentation framework and were contributed by
this research. The data processing sub-system was extended with an
AMAnnotation class to annotate OLIR pairs as arguments and output the D2
dataset described in the Dataset section. The model generation subsystem
was extended with an AMGraphBuilder class that builds and connects the
D2 dataset into a symbolic argumentation graph. The feature extraction
subsystem added the AMAligner class which derives 12 argument features
from the argumentation graphs. The classification subsystem, including
the random forest algorithm was reused with a modified, more
deterministic, pipeline orchestration that performed 10 seeded
experiments. The average per seed results were fed into the statistical
analysis subsystem which was extended to include Wilcoxon and Kendall's
W analysis, see Figure 3 for augmented AM extensions.

\begin{figure}[H]  
\centering
\includegraphics[width=0.95\linewidth,height=0.82\textheight,keepaspectratio]{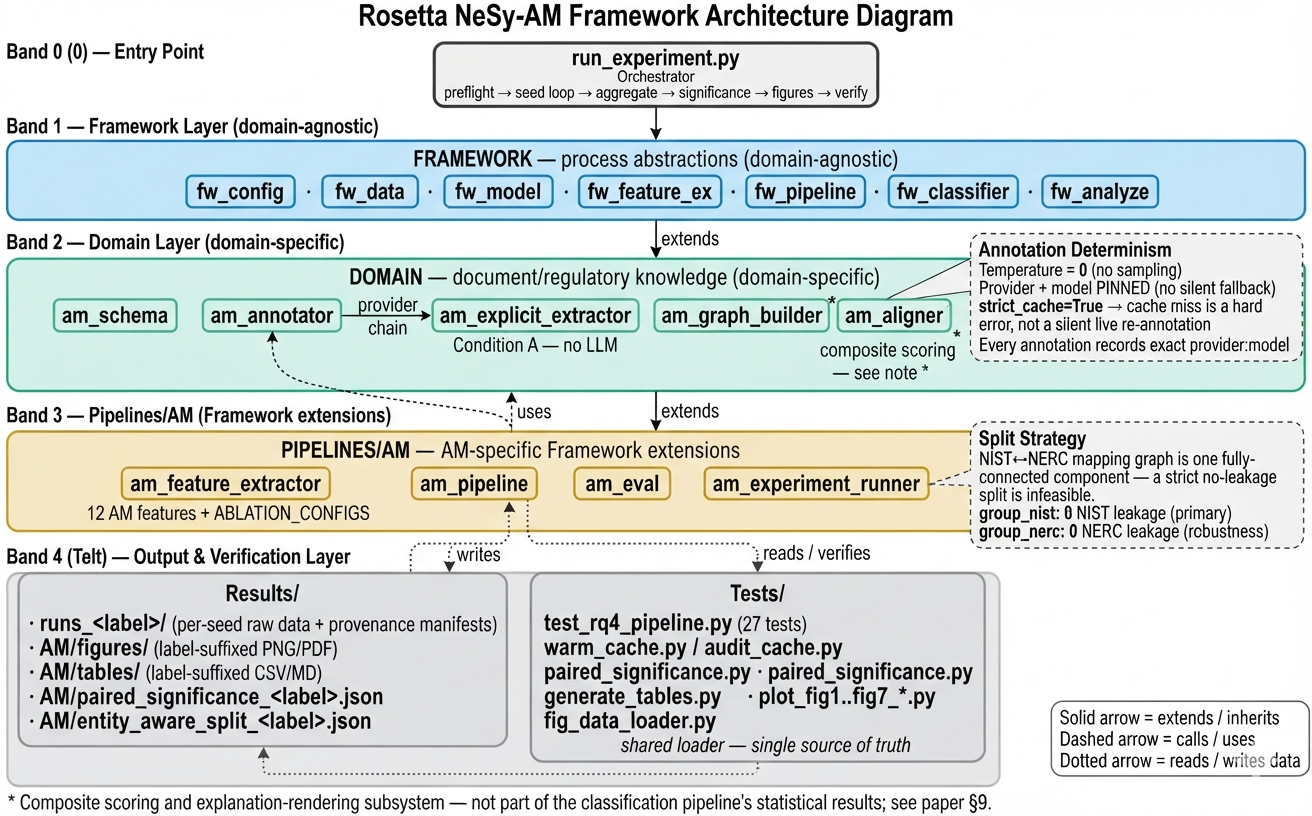}
\caption{AM System Architecture}
\label{fig:am-system-architecture}
\end{figure}

The argumentation extension system architecture, see Figure 3, shows a
more detailed separation between the core and extended capabilities into
two layers. A domain agnostic core framework layer and a domain specific
extension layer. The domain agnostic framework layer provides reusable
process abstractions (data handling, feature extraction, classification,
statistical analysis, and pipeline stage orchestration). The domain
layer encodes domain specific knowledge (e.g., regulatory and Toulmin
argumentation theory). This separation encapsulates any task or use case
specific knowledge enabling reusability and portability to other domains
without the underlying framework requiring major modifications. The
extended AM framework enables the pipelines used to generate and analyze
the experimental results of this research. The modified AM workflow is
orchestrated across the core pipelines (TF-IDF, NeSy-only, NeSy+AM) and
their variants (NeSy+AM-Jaccard, NeSy+AM-cosine, NeSy+AM-Gupta), see
Figure 4 below

\begin{figure}[H]  
\centering
\includegraphics[width=0.95\linewidth,height=0.82\textheight,keepaspectratio]{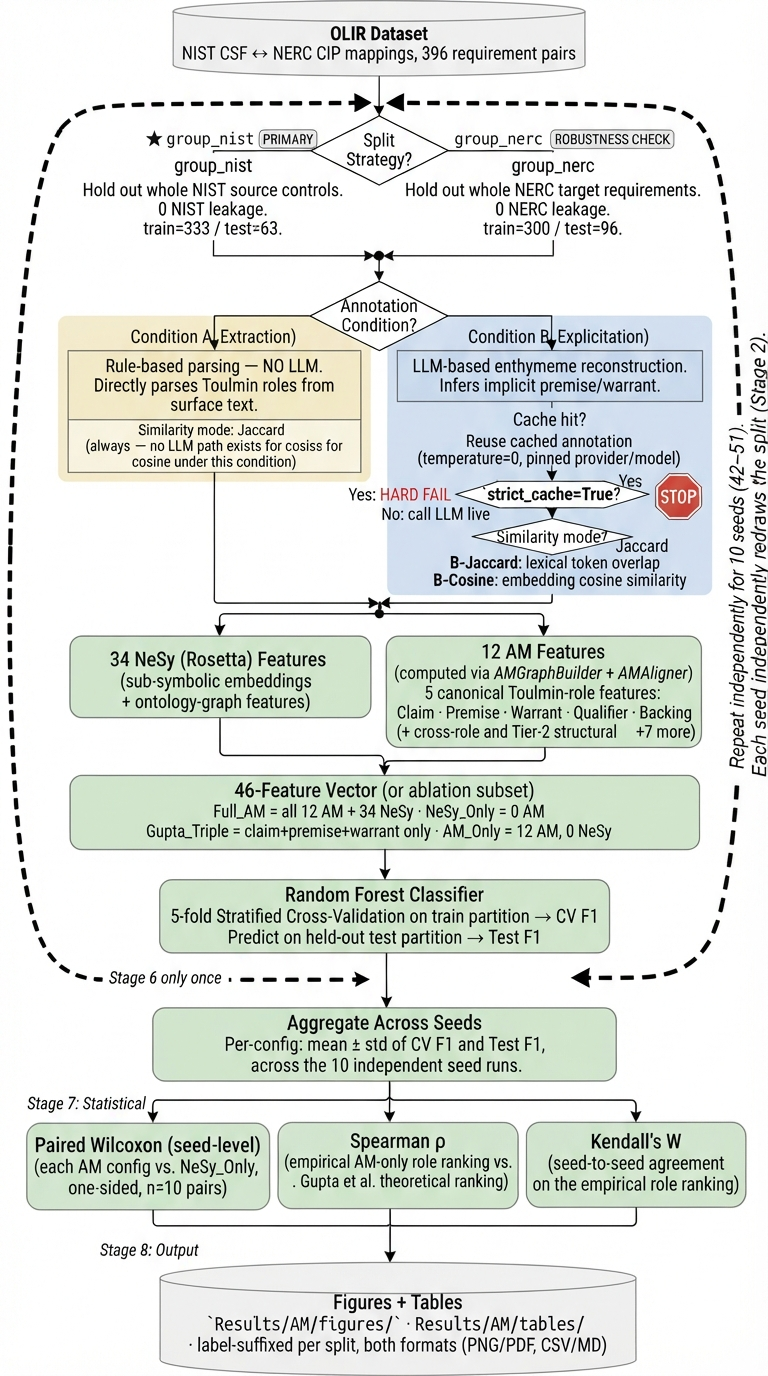}
\caption{AM Experiment Dataflow}
\label{fig:am-experiment-dataflow}
\end{figure}

\paragraph{}\label{section}

\paragraph{Feature Extraction and
Aggregation}\label{feature-extraction-and-aggregation}

The AM pipeline orchestration described in Figure 4 classifies pairs
consisting of two argument graphs, one from each standard corpus
(NIST-CSF, NERC-CIP). This is the input to the AMAligner class which
computes pairwise similarity across each Toulmin role. For roles with
multiple nodes, the Hungarian Algorithm is used to find the optimal
assignment. Specifically, for each role an n x m cosine similarity
matrix is constructed and the Greedy Hungarian Algorithm finds the
optimal assignments that maximize total intra-role similarity. Each of
the roles are computed independently resulting in five pairwise
similarity scalars representing the optimal argumentation alignment
across pairs.

The AMAligner extracts the five direct Toulmin role similarity scalars
and computes an additional seven derived engineered features. The
engineered features measure cross-role similarity, structural
relationship alignment, scope, and relationship direction. The formulas
used to compare pairs of model representations are listed in Appendix D.
These formulas are implemented as algorithms in the SciPy and NumPy
python libraries and were used to extract the AM features. Following
standard NLP practice (Peldszus and Stede, 2013),
\textquotesingle premise\textquotesingle{} is used to refer to
Toulmin\textquotesingle s grounds component, reflecting its role as the
evidential basis for the claim. The 12 total extracted argument features
based on Toulmin's argumentation are listed with their argumentation
roles and described in Table 3 below.

\begin{longtable}[]{@{}
  >{\raggedright\arraybackslash}p{(\columnwidth - 2\tabcolsep) * \real{0.3580}}
  >{\raggedright\arraybackslash}p{(\columnwidth - 2\tabcolsep) * \real{0.6420}}@{}}
\caption{Argument Classification Features}\tabularnewline
\toprule\noalign{}
\begin{minipage}[b]{\linewidth}\raggedright
Feature
\end{minipage} & \begin{minipage}[b]{\linewidth}\raggedright
Description
\end{minipage} \\
\midrule\noalign{}
\endfirsthead
\toprule\noalign{}
\begin{minipage}[b]{\linewidth}\raggedright
Feature
\end{minipage} & \begin{minipage}[b]{\linewidth}\raggedright
Description
\end{minipage} \\
\midrule\noalign{}
\endhead
\bottomrule\noalign{}
\endlastfoot
am\_claim\_cosine & Security-objective alignment between the two
requirements\textquotesingle{} claims \\
am\_premise\_similarity & Mean similarity across matched premise spans
(factual basis alignment) \\
am\_warrant\_similarity & Mean similarity across matched warrant spans
(inferential reasoning alignment) \\
am\_qualifier\_similarity & Applicability-scope similarity \\
am\_backing\_overlap & Normalized overlap in shared normative/citation
authority \\
am\_grounds\_warrant\_coherence & Premise$\leftrightarrow$warrant cross-role
coherence \\
am\_gwc\_directionality & Signed direction of the grounds-warrant
coherence relationship \\
am\_qualifier\_scope\_containment & Asymmetry in qualifier scope (does
one requirement\textquotesingle s scope contain the
other\textquotesingle s) \\
am\_backing\_citation\_asymmetry & Directional containment between the
two requirements\textquotesingle{} citations \\
am\_obligation\_tier\_match & Match on obligation strength
(shall/should/may) \\
am\_backing\_family\_overlap & Jaccard overlap of citation families \\
am\_subject\_entity\_match & Match on subject entity type
(person/system/process) \\
\end{longtable}

\paragraph{Classification}\label{classification}

Classification is performed by an open source Random Forest (RF)
algorithm provided via the Python library sklearn. The RF algorithm
learns implicit feature importance through bootstrap aggregation and
random feature subsampling at each tree split. Gini impurity importance
scores are calculated via ensemble training. The RF hyperparameters used
are specified in Table 4 below. Model selection and performance
estimation used a stratified 5-fold cross-validation on the training
partition. CV-F1 and Test-F1 are the primary metrics. The default
decision threshold of 0.5 was used.

Two techniques were used to determine individual feature importance in
optimizing F1 classification performance. The RF classifier natively
finds the optimal split thresholds across trees. At each node in the
tree, the classifier splits on the feature and threshold that
contributes the least Gini impurity from the training dataset. Therefore
the fully trained classification tree implicitly contains the Gini
importance of the training dataset. A post-hoc Gini importance is then
extracted from the tree which explicitly defines what features the RF
classifier used and their importance.

However, Gini importance has two known issues that were considered and
mitigated with a further feature permutation experiment. The first issue
is its bias toward continuous and high-cardinality features. Features
with greater unique values (continuous cosine vectors vs discrete
labels) receive more split opportunities in the RF tree and therefore
higher Gini importance regardless of actual predictive value. The second
Gini bias is that it is calculated on training data only and therefore
is a measurement of how important the feature was in fitting the
training dataset.

Therefore, a separate permutation importance post-hoc test isolating
each feature was completed. Permutation importance is computed on the
test dataset and measures the actual feature importance to a generalized
model. Each AM feature's `column' in the test set is randomly shuffled
which breaks the relationship between the feature and the OLIR
harmonization label. The mean decrease in F1 repeated scores (n\_repeats
= 30) after the RF classifier was run with the randomized feature column
reflects the features permutation importance (Altmann et al., 2010).

\begin{longtable}[]{@{}
  >{\raggedright\arraybackslash}p{(\columnwidth - 2\tabcolsep) * \real{0.3365}}
  >{\raggedright\arraybackslash}p{(\columnwidth - 2\tabcolsep) * \real{0.6635}}@{}}
\caption{Hyperparameters}\tabularnewline
\toprule\noalign{}
\begin{minipage}[b]{\linewidth}\raggedright
Hyperparameter
\end{minipage} & \begin{minipage}[b]{\linewidth}\raggedright
Value
\end{minipage} \\
\midrule\noalign{}
\endfirsthead
\toprule\noalign{}
\begin{minipage}[b]{\linewidth}\raggedright
Hyperparameter
\end{minipage} & \begin{minipage}[b]{\linewidth}\raggedright
Value
\end{minipage} \\
\midrule\noalign{}
\endhead
\bottomrule\noalign{}
\endlastfoot
source & sklearn.ensemble.RandomForestClassifier \\
n\_estimators & 100 \\
max\_depth & None \\
class\_weight & Balanced \\
max\_features & Sqrt \\
random\_state & 42 \\
\end{longtable}

\paragraph{Evaluation Metrics and Statistical
Analysis}\label{evaluation-metrics-and-statistical-analysis}

\subparagraph{F1-CV and F1-test Metric}\label{f1-cv-and-f1-test-metric}

The primary performance metric used across experiments is F1 which
matches the harmonization use case success criteria, see equations in
Appendix D. In standards harmonization, both false positives
(classifying non-harmonizable as harmonizable) and false negatives
(missing harmonizable classifications) waste resources and increase
costs. These scenarios map directly to precision and recall, showing
that neither in isolation is the correct performance metric, but instead
their balance or F1 score is the optimal performance metric.

\subparagraph{\texorpdfstring{Wilcoxon Signed-Rank
}{Wilcoxon Signed-Rank }}\label{wilcoxon-signed-rank}

Each research question essentially compares the F1 performance or the
ranking order of the results between different models. These results are
statistically analyzed to determine if their performance differences are
random noise in the system or if they are statistically significant. In
all scenarios the normality of the scores cannot be assumed so
non-parametric analyses are required. Wilcoxon signed rank (and Cohen's
$d_z$, its companion effect size statistic) and Kendall's W are the
statistical analysis appropriate for each of their respective research
question and are described below and in Appendix D.

Wilcoxon is used to analyze the directional hypothesis of RQ1 and RQ3
(do the F1 performance results of one model significant outperform
another model). It tests the significance between different models F1
scores. It is a signed-rank, standard, nonparametric paired/dependent
statistical test that makes no assumptions on the normality of the
dataset. The open source Python library scipy.stats was used as the
implementation for all Wilcoxon statistical tests (Virtanen et al.,
2020).

The null hypothesis ($H_0$) for the Wilcoxon test is the median of the
paired differences will be zero indicating paired samples were drawn
from the same distribution. To test $H_0$, 10 independently seeded pipeline
runs are completed that each redraw the configured group based split
(NERC or NIST, depending on the analysis see Split Methodology section)
to yield one 5-fold cross validation mean and one held out test F1
score. The Wilcoxon test then pairs the 10 seed level mean CV-F1 scores
between models (seed is the unit of independent replication). A p-value
for the median of the differences between the CV-F1 scores from the
models under comparison is computed to determine if the difference
between the models is statistically significant from zero ($H_0$)
(Hollander \& Wolfe, 2014).

In RQ2, the Wilcoxon signed-rank test analyzed if different AM
annotation methodologies influence F1 classification performance.
Additionally, in RQ1 it was used to analyze the primary research
question, if the additional AM features and specific subsets of the AM
features improve the F1 classification performance. These are
directional hypotheses and therefore used directional one-sided Wilcoxon
tests.

\subparagraph{Cohen's $d_z$}\label{cohens-$d_z$} 

RQ1 and RQ3, in addition to the Wilcoxon significance test, described
separately in the Evaluation Metrics section, used its companion effect
size metric, Cohen's $d_z$ measures the standardized mean difference
between two related/paired groups. It signifies how large the difference
is in standard deviation units, using the standard deviation of the
differences (not the pooled SD). Cohen's $d_z$, rank-biserial
r, and confidence intervals (CI 95\%) for each comparison were
calculated and the results described in the Results section.

\subparagraph{Kendall's W}\label{kendalls-w}

RQ3 compared the ranking order of feature importance across
configurations and experimental seeds. The focus of the analysis was the
reliability of the empirical rankings across seeds (10). Kendall's W
measures the stability of the relative ranking of features across seeds.
It is tie-corrected, chi-square test that with 10 seeds is adequately
powered to analyze the reliability dimension of RQ3.

\subsection{\texorpdfstring{Experiment Validation
}{Experiment Validation }}\label{experiment-validation}

\paragraph{\texorpdfstring{Mirror Group Data-Split Strategy
}{Mirror Group Data-Split Strategy }}\label{mirror-group-data-split-strategy}

To validate the robustness of the experimental results and mitigate the
data leakage risks inherent in the OLIR dataset described in the Dataset
section, a mirror group data-split strategy was added to the experiment.
As described, the OLIR dataset has a many to many relationship structure
between NIST and NERC mappings In pairwise dataset construction, this
creates the potential for information leakage across the initial
training/CV and test splits.

This inherent quality of the dataset means that in a randomly selected
data-split, a NIST-CSF requirement could be mapped to multiple NERC-CIP
requirements and the reverse is also true. Standard practice is to
mitigate this issue via entity-data splits or group-data splits. The
deeply nested structure of the dataset makes pure entity-aware splits,
where references are considered a single entity and are not split
between partitions infeasible. The OLIR mapping graph would reduce to a
single fully connected component under transitive closure, making a
strict entity-aware split result in an empty partition. Therefore, a
group-data split that holds out entire source controls was implemented
for experiment validation.

This strategy holds out entire controls (NIST held out) so there is 0
data leakage of controls across training and test splits. This research
also added additional validation robustness by adding a mirrored
experiment (NERC held out). The primary experiment uses group-by-NIST
split strategy which assures 0 NIST control data leakage and tests the
actual OLIR deployment scenario that, given a new NIST control, predict
the mapping to a known NERC corpus. This measures generalization to
unseen NIST source controls against a known targeted corpus. While the
group-by-NIST holds out the source control cleanly and addresses the
deployment scenario, from a data science perspective it also means the
NERC requirements can overlap (empirically in the experimental splits
79\% (42/53) of the NERC requirements overlapped). For this reason, the
additional mirrored data split experiment was also run. The mirrored
group-by-NERC split strategy is a completely separate, mutually
exclusive splitting strategy.

In summary, both a robustness check within the group-by-NIST test set
was done on the fully disjointed NERC subset (11/53 NERC requirements)
and a complete mirrored group-by-NERC split strategy was done as an
independent robustness validation. Each split strategy holds out one
side of the data (NERC or NIST). Both (group-by-NIST and group-by-NERC)
run independently over 10 seeded experiments. Group-by-NIST is the
primary comparison. Group-by-NERC is the robustness validation test.

\paragraph{\texorpdfstring{Reproducibility
}{Reproducibility }}\label{reproducibility}

All experiments are deterministic given the 10 fixed seeds (42-51),
applied independently per run to the group-based split, RF, and
within-train CV fold assignments. Additionally, a consistent group based
train/test split on the held-out group (group\_NIST; 333/63,
group\_NERC; 300/96) was used across experiments. Finally, the LLM
annotation was pinned to a model, version and temperature set to be
deterministic (Llama-3.3-70B-Versatile via Groq, temperature=0, cached
in persistent JSON store). The full replication package includes dataset
construction code, annotation cache, feature extraction pipeline,
classifier configuration, and all random seeds and is available upon
request.

\section{Results and Analysis}\label{results-and-analysis}

This research is based on the theory that in domains which are organized
around explicit or implicit reasoning, leveraging argumentation mining
(AM) techniques yields richer semantic representations. From this theory
we hypothesize that cybersecurity standards can be modeled as formal
Toulmin argumentation structures to improve the classification accuracy
of harmonizing NERC-CIP and NIST-CSF frameworks. To investigate this
hypothesis, an NLP pipeline framework and a set of experiments were
developed to empirically test RQ1-3. All experiments used the full
10-seed experimental run and the OLIR dataset with the NIST
split-strategy (group\_NIST, train=333, test = 63, 0 NIST leakage) and
were validated using the mirror NERC split-strategy (group\_NERC,
train=300, test=96, 0 NERC leakage), as described in the Experimental
Setup and Evaluation Protocol section.

\subsection{RQ1 -- Does Argument Structure Improve Harmonization
Performance?}\label{rq1-does-argument-structure-improve-harmonization-performance}

Argument structure improves harmonization performance. The argument
aware models consistently outperformed the neuro-symbolic semantic
baseline across the repeated cross validation experiments. The
improvement Is also observed on the held out control level test set. The
result provides evidence supporting the central hypothesis: argument
derived representations contain information useful for normative text
alignment beyond the information captured by the baseline semantic
representation.

RQ1 asks, what is the effect of integrating Toulmin-derived argument
mining features into a neuro-symbolic NLP pipeline for NERC-CIP and
NIST-CSF standards harmonization. To investigate RQ1, three new
integrated NeSy+AM models were developed and tested. The three models:
NeSy+AM-C, NeSy+AM-J, NeSy+AM-Gupta triple, were run through the full
10-seed 5-fold CV NLP classification experiment. Their results were
compared with the base NeSy and TF-IDF models. The combined five model
comparison and statistical analysis is described below in Tables 5-6 and
summarized in Figure 5 below.

\begin{longtable}[]{@{}
  >{\raggedright\arraybackslash}p{(\columnwidth - 12\tabcolsep) * \real{0.1546}}
  >{\raggedright\arraybackslash}p{(\columnwidth - 12\tabcolsep) * \real{0.2803}}
  >{\raggedright\arraybackslash}p{(\columnwidth - 12\tabcolsep) * \real{0.1220}}
  >{\raggedright\arraybackslash}p{(\columnwidth - 12\tabcolsep) * \real{0.1029}}
  >{\raggedright\arraybackslash}p{(\columnwidth - 12\tabcolsep) * \real{0.1335}}
  >{\raggedright\arraybackslash}p{(\columnwidth - 12\tabcolsep) * \real{0.1145}}
  >{\raggedright\arraybackslash}p{(\columnwidth - 12\tabcolsep) * \real{0.0923}}@{}}
\caption{Pipeline F1-CV and F1-Test Scores}\tabularnewline
\toprule\noalign{}
\begin{minipage}[b]{\linewidth}\raggedright
Condition
\end{minipage} & \begin{minipage}[b]{\linewidth}\raggedright
config
\end{minipage} & \begin{minipage}[b]{\linewidth}\raggedright
cv\_f1\_mean
\end{minipage} & \begin{minipage}[b]{\linewidth}\raggedright
cv\_f1\_std
\end{minipage} & \begin{minipage}[b]{\linewidth}\raggedright
test\_f1\_mean
\end{minipage} & \begin{minipage}[b]{\linewidth}\raggedright
test\_f1\_std
\end{minipage} & \begin{minipage}[b]{\linewidth}\raggedright
n\_seeds
\end{minipage} \\
\midrule\noalign{}
\endfirsthead
\toprule\noalign{}
\begin{minipage}[b]{\linewidth}\raggedright
Condition
\end{minipage} & \begin{minipage}[b]{\linewidth}\raggedright
config
\end{minipage} & \begin{minipage}[b]{\linewidth}\raggedright
cv\_f1\_mean
\end{minipage} & \begin{minipage}[b]{\linewidth}\raggedright
cv\_f1\_std
\end{minipage} & \begin{minipage}[b]{\linewidth}\raggedright
test\_f1\_mean
\end{minipage} & \begin{minipage}[b]{\linewidth}\raggedright
test\_f1\_std
\end{minipage} & \begin{minipage}[b]{\linewidth}\raggedright
n\_seeds
\end{minipage} \\
\midrule\noalign{}
\endhead
\bottomrule\noalign{}
\endlastfoot
n/a & NeSy Only (prior work) & 0.7473 & 0.0252 & 0.7375 & 0.0407 & 10 \\
B-Jaccard & NeSy + Full AM (12 features) & 0.7641 & 0.0253 & 0.7438 &
0.0449 & 10 \\
B-Jaccard & NeSy + Gupta Triple (claim/premise/warrant) & 0.7618 &
0.0198 & 0.7528 & 0.0404 & 10 \\
B-Jaccard & AM Only (no NeSy base) & 0.6658 & 0.0307 & 0.6389 & 0.0541 &
10 \\
B-Cosine & NeSy + Full AM (12 features) & 0.7786 & 0.0202 & 0.7895 &
0.0615 & 10 \\
B-Cosine & NeSy + Gupta Triple (claim/premise/warrant) & 0.7712 & 0.0252
& 0.787 & 0.0563 & 10 \\
B-Cosine & AM Only (no NeSy base) & 0.7736 & 0.0218 & 0.7313 & 0.0424 &
10 \\
\end{longtable}

The results of the primary comparisons (NeSy+AM-C/NeSy-only,
NeSy+AM-Gupta/NeSy-only) showed a statistically significant
classification performance improvement using the fused NeSy+AM models. A
paired Wilcoxon signed rank test between the NeSy+AM-cosine and
NeSy-only models yielded a CV-F1 $\Delta\approx$+0.031, p=0.0049 and Test-F1, $\Delta=$
+0.052, p=0.019, with a large effect size (Cohen's $d_z$=0.817). Variance
across seeds and folds clustered around the means ($\sigma=$0.020-0.025), see
Table 6 below.

In contrast the results across the fused models under Jaccard conditions
were statistically significant only under the CV-F1 metric at an
uncorrected $\alpha=$0.05. Classification performance of the non-fused
AM-feature NeSy-only model followed the same pattern as the Jaccard
models and were not significantly different from the NeSy only model
under Test-F1, see Table 6 below.

These experimental results considered together, suggest AM features
derived from explicitly annotated Toulmin models improve classification
performance that generalizes across CV and held out Test-F1.
Specifically, that (a) there is genuine signal carried in the
argumentation features and (b) the classification benefit is in the
fusion of the NeSy+AM features rather than the features in isolation, as
summarized in Figure 5 below.

\clearpage\begin{landscape}
{\scriptsize\setlength{\tabcolsep}{2pt}\renewcommand{\arraystretch}{0.92}
\begin{longtable}[]{@{}
  >{\raggedright\arraybackslash}p{(\columnwidth - 42\tabcolsep) * \real{0.0757}}
  >{\raggedright\arraybackslash}p{(\columnwidth - 42\tabcolsep) * \real{0.1054}}
  >{\raggedright\arraybackslash}p{(\columnwidth - 42\tabcolsep) * \real{0.0667}}
  >{\raggedright\arraybackslash}p{(\columnwidth - 42\tabcolsep) * \real{0.0477}}
  >{\raggedright\arraybackslash}p{(\columnwidth - 42\tabcolsep) * \real{0.0381}}
  >{\raggedright\arraybackslash}p{(\columnwidth - 42\tabcolsep) * \real{0.0667}}
  >{\raggedright\arraybackslash}p{(\columnwidth - 42\tabcolsep) * \real{0.0667}}
  >{\raggedright\arraybackslash}p{(\columnwidth - 42\tabcolsep) * \real{0.0667}}
  >{\raggedright\arraybackslash}p{(\columnwidth - 42\tabcolsep) * \real{0.0667}}
  >{\raggedright\arraybackslash}p{(\columnwidth - 42\tabcolsep) * \real{0.0763}}
  >{\raggedright\arraybackslash}p{(\columnwidth - 42\tabcolsep) * \real{0.0274}}
  >{\raggedright\arraybackslash}p{(\columnwidth - 42\tabcolsep) * \real{0.0250}}
  >{\raggedright\arraybackslash}p{(\columnwidth - 42\tabcolsep) * \real{0.0250}}
  >{\raggedright\arraybackslash}p{(\columnwidth - 42\tabcolsep) * \real{0.0093}}
  >{\raggedright\arraybackslash}p{(\columnwidth - 42\tabcolsep) * \real{0.0497}}
  >{\raggedright\arraybackslash}p{(\columnwidth - 42\tabcolsep) * \real{0.0093}}
  >{\raggedright\arraybackslash}p{(\columnwidth - 42\tabcolsep) * \real{0.0489}}
  >{\raggedright\arraybackslash}p{(\columnwidth - 42\tabcolsep) * \real{0.0093}}
  >{\raggedright\arraybackslash}p{(\columnwidth - 42\tabcolsep) * \real{0.0625}}
  >{\raggedright\arraybackslash}p{(\columnwidth - 42\tabcolsep) * \real{0.0093}}
  >{\raggedright\arraybackslash}p{(\columnwidth - 42\tabcolsep) * \real{0.0378}}
  >{\raggedright\arraybackslash}p{(\columnwidth - 42\tabcolsep) * \real{0.0093}}@{}}
\caption{Pipeline Statistical Analysis}\tabularnewline
\toprule\noalign{}
\begin{minipage}[b]{\linewidth}\raggedright
Cond.
\end{minipage} & \begin{minipage}[b]{\linewidth}\raggedright
config
\end{minipage} & \begin{minipage}[b]{\linewidth}\raggedright
base
\end{minipage} & \begin{minipage}[b]{\linewidth}\raggedright
Metric
\end{minipage} & \begin{minipage}[b]{\linewidth}\raggedright
n
\end{minipage} & \begin{minipage}[b]{\linewidth}\raggedright
Base mean
\end{minipage} & \begin{minipage}[b]{\linewidth}\raggedright
Am mean
\end{minipage} & \begin{minipage}[b]{\linewidth}\raggedright
Mean delta
\end{minipage} & \begin{minipage}[b]{\linewidth}\raggedright
delta\_ci95\_low
\end{minipage} & \begin{minipage}[b]{\linewidth}\raggedright
delta\_ci95\_high
\end{minipage} & \begin{minipage}[b]{\linewidth}\raggedright
w
\end{minipage} & \begin{minipage}[b]{\linewidth}\raggedright
l
\end{minipage} & \begin{minipage}[b]{\linewidth}\raggedright
t
\end{minipage} &
\multicolumn{2}{>{\raggedright\arraybackslash}p{(\columnwidth - 42\tabcolsep) * \real{0.0590} + 2\tabcolsep}}{%
\begin{minipage}[b]{\linewidth}\raggedright
wilcoxon\_p
\end{minipage}} &
\multicolumn{2}{>{\raggedright\arraybackslash}p{(\columnwidth - 42\tabcolsep) * \real{0.0583} + 2\tabcolsep}}{%
\begin{minipage}[b]{\linewidth}\raggedright
cohens\_dz
\end{minipage}} &
\multicolumn{2}{>{\raggedright\arraybackslash}p{(\columnwidth - 42\tabcolsep) * \real{0.0718} + 2\tabcolsep}}{%
\begin{minipage}[b]{\linewidth}\raggedright
rank\_biserial\_r
\end{minipage}} &
\multicolumn{2}{>{\raggedright\arraybackslash}p{(\columnwidth - 42\tabcolsep) * \real{0.0471} + 2\tabcolsep}}{%
\begin{minipage}[b]{\linewidth}\raggedright
sig
\end{minipage}} & \begin{minipage}[b]{\linewidth}\raggedright
\end{minipage} \\
\midrule\noalign{}
\endfirsthead
\toprule\noalign{}
\begin{minipage}[b]{\linewidth}\raggedright
Cond.
\end{minipage} & \begin{minipage}[b]{\linewidth}\raggedright
config
\end{minipage} & \begin{minipage}[b]{\linewidth}\raggedright
base
\end{minipage} & \begin{minipage}[b]{\linewidth}\raggedright
Metric
\end{minipage} & \begin{minipage}[b]{\linewidth}\raggedright
n
\end{minipage} & \begin{minipage}[b]{\linewidth}\raggedright
Base mean
\end{minipage} & \begin{minipage}[b]{\linewidth}\raggedright
Am mean
\end{minipage} & \begin{minipage}[b]{\linewidth}\raggedright
Mean delta
\end{minipage} & \begin{minipage}[b]{\linewidth}\raggedright
delta\_ci95\_low
\end{minipage} & \begin{minipage}[b]{\linewidth}\raggedright
delta\_ci95\_high
\end{minipage} & \begin{minipage}[b]{\linewidth}\raggedright
w
\end{minipage} & \begin{minipage}[b]{\linewidth}\raggedright
l
\end{minipage} & \begin{minipage}[b]{\linewidth}\raggedright
t
\end{minipage} &
\multicolumn{2}{>{\raggedright\arraybackslash}p{(\columnwidth - 42\tabcolsep) * \real{0.0590} + 2\tabcolsep}}{%
\begin{minipage}[b]{\linewidth}\raggedright
wilcoxon\_p
\end{minipage}} &
\multicolumn{2}{>{\raggedright\arraybackslash}p{(\columnwidth - 42\tabcolsep) * \real{0.0583} + 2\tabcolsep}}{%
\begin{minipage}[b]{\linewidth}\raggedright
cohens\_dz
\end{minipage}} &
\multicolumn{2}{>{\raggedright\arraybackslash}p{(\columnwidth - 42\tabcolsep) * \real{0.0718} + 2\tabcolsep}}{%
\begin{minipage}[b]{\linewidth}\raggedright
rank\_biserial\_r
\end{minipage}} &
\multicolumn{2}{>{\raggedright\arraybackslash}p{(\columnwidth - 42\tabcolsep) * \real{0.0471} + 2\tabcolsep}}{%
\begin{minipage}[b]{\linewidth}\raggedright
sig
\end{minipage}} & \begin{minipage}[b]{\linewidth}\raggedright
\end{minipage} \\
\midrule\noalign{}
\endhead
\bottomrule\noalign{}
\endlastfoot
B-Jaccard & NeSy + Full AM (12 features) & NeSy\_Only & Cv & 10 & 0.7473
& 0.7641 & 0.0168 & 0.0075 & 0.0281 & 8 & 2 &
\multicolumn{2}{>{\raggedright\arraybackslash}p{(\columnwidth - 42\tabcolsep) * \real{0.0343} + 2\tabcolsep}}{%
0} &
\multicolumn{2}{>{\raggedright\arraybackslash}p{(\columnwidth - 42\tabcolsep) * \real{0.0590} + 2\tabcolsep}}{%
0.0049} &
\multicolumn{2}{>{\raggedright\arraybackslash}p{(\columnwidth - 42\tabcolsep) * \real{0.0583} + 2\tabcolsep}}{%
0.953} &
\multicolumn{2}{>{\raggedright\arraybackslash}p{(\columnwidth - 42\tabcolsep) * \real{0.0718} + 2\tabcolsep}}{%
0.891} &
\multicolumn{2}{>{\raggedright\arraybackslash}p{(\columnwidth - 42\tabcolsep) * \real{0.0471} + 2\tabcolsep}@{}}{%
*} \\
B-Jaccard & NeSy + Full AM (12 features) & NeSy\_Only & test & 10 &
0.7375 & 0.7438 & 0.0063 & -0.0169 & 0.0303 & 6 & 4 &
\multicolumn{2}{>{\raggedright\arraybackslash}p{(\columnwidth - 42\tabcolsep) * \real{0.0343} + 2\tabcolsep}}{%
0} &
\multicolumn{2}{>{\raggedright\arraybackslash}p{(\columnwidth - 42\tabcolsep) * \real{0.0590} + 2\tabcolsep}}{%
0.3125} &
\multicolumn{2}{>{\raggedright\arraybackslash}p{(\columnwidth - 42\tabcolsep) * \real{0.0583} + 2\tabcolsep}}{%
0.155} &
\multicolumn{2}{>{\raggedright\arraybackslash}p{(\columnwidth - 42\tabcolsep) * \real{0.0718} + 2\tabcolsep}}{%
0.2} &
\multicolumn{2}{>{\raggedright\arraybackslash}p{(\columnwidth - 42\tabcolsep) * \real{0.0471} + 2\tabcolsep}@{}}{%
} \\
B-Jaccard & NeSy + Gupta Triple (claim/premise/warrant) & NeSy\_Only &
cv & 10 & 0.7473 & 0.7618 & 0.0145 & 0.003 & 0.0275 & 7 & 3 &
\multicolumn{2}{>{\raggedright\arraybackslash}p{(\columnwidth - 42\tabcolsep) * \real{0.0343} + 2\tabcolsep}}{%
0} &
\multicolumn{2}{>{\raggedright\arraybackslash}p{(\columnwidth - 42\tabcolsep) * \real{0.0590} + 2\tabcolsep}}{%
0.0322} &
\multicolumn{2}{>{\raggedright\arraybackslash}p{(\columnwidth - 42\tabcolsep) * \real{0.0583} + 2\tabcolsep}}{%
0.688} &
\multicolumn{2}{>{\raggedright\arraybackslash}p{(\columnwidth - 42\tabcolsep) * \real{0.0718} + 2\tabcolsep}}{%
0.673} &
\multicolumn{2}{>{\raggedright\arraybackslash}p{(\columnwidth - 42\tabcolsep) * \real{0.0471} + 2\tabcolsep}@{}}{%
*} \\
B-Jaccard & NeSy + Gupta Triple (claim/premise/warrant) & NeSy\_Only &
test & 10 & 0.7375 & 0.7528 & 0.0153 & -0.0081 & 0.0423 & 4 & 4 &
\multicolumn{2}{>{\raggedright\arraybackslash}p{(\columnwidth - 42\tabcolsep) * \real{0.0343} + 2\tabcolsep}}{%
2} &
\multicolumn{2}{>{\raggedright\arraybackslash}p{(\columnwidth - 42\tabcolsep) * \real{0.0590} + 2\tabcolsep}}{%
0.2734} &
\multicolumn{2}{>{\raggedright\arraybackslash}p{(\columnwidth - 42\tabcolsep) * \real{0.0583} + 2\tabcolsep}}{%
0.354} &
\multicolumn{2}{>{\raggedright\arraybackslash}p{(\columnwidth - 42\tabcolsep) * \real{0.0718} + 2\tabcolsep}}{%
0.278} &
\multicolumn{2}{>{\raggedright\arraybackslash}p{(\columnwidth - 42\tabcolsep) * \real{0.0471} + 2\tabcolsep}@{}}{%
} \\
B-Jaccard & AM Only (no NeSy base) & NeSy\_Only & cv & 10 & 0.7473 &
0.6658 & -0.0815 & -0.1055 & -0.059 & 0 & 10 &
\multicolumn{2}{>{\raggedright\arraybackslash}p{(\columnwidth - 42\tabcolsep) * \real{0.0343} + 2\tabcolsep}}{%
0} &
\multicolumn{2}{>{\raggedright\arraybackslash}p{(\columnwidth - 42\tabcolsep) * \real{0.0590} + 2\tabcolsep}}{%
1.0} &
\multicolumn{2}{>{\raggedright\arraybackslash}p{(\columnwidth - 42\tabcolsep) * \real{0.0583} + 2\tabcolsep}}{%
-2.048} &
\multicolumn{2}{>{\raggedright\arraybackslash}p{(\columnwidth - 42\tabcolsep) * \real{0.0718} + 2\tabcolsep}}{%
-1.0} &
\multicolumn{2}{>{\raggedright\arraybackslash}p{(\columnwidth - 42\tabcolsep) * \real{0.0471} + 2\tabcolsep}@{}}{%
} \\
B-Jaccard & AM Only (no NeSy base) & NeSy\_Only & test & 10 & 0.7375 &
0.6389 & -0.0986 & -0.1488 & -0.05 & 2 & 8 &
\multicolumn{2}{>{\raggedright\arraybackslash}p{(\columnwidth - 42\tabcolsep) * \real{0.0343} + 2\tabcolsep}}{%
0} &
\multicolumn{2}{>{\raggedright\arraybackslash}p{(\columnwidth - 42\tabcolsep) * \real{0.0590} + 2\tabcolsep}}{%
0.9971} &
\multicolumn{2}{>{\raggedright\arraybackslash}p{(\columnwidth - 42\tabcolsep) * \real{0.0583} + 2\tabcolsep}}{%
-1.17} &
\multicolumn{2}{>{\raggedright\arraybackslash}p{(\columnwidth - 42\tabcolsep) * \real{0.0718} + 2\tabcolsep}}{%
-0.891} &
\multicolumn{2}{>{\raggedright\arraybackslash}p{(\columnwidth - 42\tabcolsep) * \real{0.0471} + 2\tabcolsep}@{}}{%
} \\
B-Cosine & NeSy + Full AM (12 features) & NeSy\_Only & cv & 10 & 0.7473
& 0.7786 & 0.0313 & 0.0166 & 0.0462 & 8 & 2 &
\multicolumn{2}{>{\raggedright\arraybackslash}p{(\columnwidth - 42\tabcolsep) * \real{0.0343} + 2\tabcolsep}}{%
0} &
\multicolumn{2}{>{\raggedright\arraybackslash}p{(\columnwidth - 42\tabcolsep) * \real{0.0590} + 2\tabcolsep}}{%
0.0049} &
\multicolumn{2}{>{\raggedright\arraybackslash}p{(\columnwidth - 42\tabcolsep) * \real{0.0583} + 2\tabcolsep}}{%
1.245} &
\multicolumn{2}{>{\raggedright\arraybackslash}p{(\columnwidth - 42\tabcolsep) * \real{0.0718} + 2\tabcolsep}}{%
0.891} &
\multicolumn{2}{>{\raggedright\arraybackslash}p{(\columnwidth - 42\tabcolsep) * \real{0.0471} + 2\tabcolsep}@{}}{%
*} \\
B-Cosine & NeSy + Full AM (12 features) & NeSy\_Only & test & 10 &
0.7375 & 0.7895 & 0.0519 & 0.0159 & 0.0902 & 8 & 2 &
\multicolumn{2}{>{\raggedright\arraybackslash}p{(\columnwidth - 42\tabcolsep) * \real{0.0343} + 2\tabcolsep}}{%
0} &
\multicolumn{2}{>{\raggedright\arraybackslash}p{(\columnwidth - 42\tabcolsep) * \real{0.0590} + 2\tabcolsep}}{%
0.0186} &
\multicolumn{2}{>{\raggedright\arraybackslash}p{(\columnwidth - 42\tabcolsep) * \real{0.0583} + 2\tabcolsep}}{%
0.817} &
\multicolumn{2}{>{\raggedright\arraybackslash}p{(\columnwidth - 42\tabcolsep) * \real{0.0718} + 2\tabcolsep}}{%
0.745} &
\multicolumn{2}{>{\raggedright\arraybackslash}p{(\columnwidth - 42\tabcolsep) * \real{0.0471} + 2\tabcolsep}@{}}{%
*} \\
B-Cosine & NeSy + Gupta Triple (claim/premise/warrant) & NeSy\_Only & cv
& 10 & 0.7473 & 0.7712 & 0.0239 & 0.0112 & 0.0363 & 9 & 1 &
\multicolumn{2}{>{\raggedright\arraybackslash}p{(\columnwidth - 42\tabcolsep) * \real{0.0343} + 2\tabcolsep}}{%
0} &
\multicolumn{2}{>{\raggedright\arraybackslash}p{(\columnwidth - 42\tabcolsep) * \real{0.0590} + 2\tabcolsep}}{%
0.0068} &
\multicolumn{2}{>{\raggedright\arraybackslash}p{(\columnwidth - 42\tabcolsep) * \real{0.0583} + 2\tabcolsep}}{%
1.121} &
\multicolumn{2}{>{\raggedright\arraybackslash}p{(\columnwidth - 42\tabcolsep) * \real{0.0718} + 2\tabcolsep}}{%
0.855} &
\multicolumn{2}{>{\raggedright\arraybackslash}p{(\columnwidth - 42\tabcolsep) * \real{0.0471} + 2\tabcolsep}@{}}{%
*} \\
B-Cosine & NeSy + Gupta Triple (claim/premise/warrant) & NeSy\_Only &
test & 10 & 0.7375 & 0.787 & 0.0494 & 0.0249 & 0.072 & 9 & 1 &
\multicolumn{2}{>{\raggedright\arraybackslash}p{(\columnwidth - 42\tabcolsep) * \real{0.0343} + 2\tabcolsep}}{%
0} &
\multicolumn{2}{>{\raggedright\arraybackslash}p{(\columnwidth - 42\tabcolsep) * \real{0.0590} + 2\tabcolsep}}{%
0.0049} &
\multicolumn{2}{>{\raggedright\arraybackslash}p{(\columnwidth - 42\tabcolsep) * \real{0.0583} + 2\tabcolsep}}{%
1.235} &
\multicolumn{2}{>{\raggedright\arraybackslash}p{(\columnwidth - 42\tabcolsep) * \real{0.0718} + 2\tabcolsep}}{%
0.891} &
\multicolumn{2}{>{\raggedright\arraybackslash}p{(\columnwidth - 42\tabcolsep) * \real{0.0471} + 2\tabcolsep}@{}}{%
*} \\
B-Cosine & AM Only (no NeSy base) & NeSy\_Only & cv & 10 & 0.7473 &
0.7736 & 0.0263 & 0.0147 & 0.0382 & 9 & 1 &
\multicolumn{2}{>{\raggedright\arraybackslash}p{(\columnwidth - 42\tabcolsep) * \real{0.0343} + 2\tabcolsep}}{%
0} &
\multicolumn{2}{>{\raggedright\arraybackslash}p{(\columnwidth - 42\tabcolsep) * \real{0.0590} + 2\tabcolsep}}{%
0.002} &
\multicolumn{2}{>{\raggedright\arraybackslash}p{(\columnwidth - 42\tabcolsep) * \real{0.0583} + 2\tabcolsep}}{%
1.3} &
\multicolumn{2}{>{\raggedright\arraybackslash}p{(\columnwidth - 42\tabcolsep) * \real{0.0718} + 2\tabcolsep}}{%
0.964} &
\multicolumn{2}{>{\raggedright\arraybackslash}p{(\columnwidth - 42\tabcolsep) * \real{0.0471} + 2\tabcolsep}@{}}{%
*} \\
B-Cosine & AM Only (no NeSy base) & NeSy\_Only & test & 10 & 0.7375 &
0.7313 & -0.0063 & -0.0336 & 0.0225 & 3 & 6 &
\multicolumn{2}{>{\raggedright\arraybackslash}p{(\columnwidth - 42\tabcolsep) * \real{0.0343} + 2\tabcolsep}}{%
1} &
\multicolumn{2}{>{\raggedright\arraybackslash}p{(\columnwidth - 42\tabcolsep) * \real{0.0590} + 2\tabcolsep}}{%
0.6738} &
\multicolumn{2}{>{\raggedright\arraybackslash}p{(\columnwidth - 42\tabcolsep) * \real{0.0583} + 2\tabcolsep}}{%
-0.13} &
\multicolumn{2}{>{\raggedright\arraybackslash}p{(\columnwidth - 42\tabcolsep) * \real{0.0718} + 2\tabcolsep}}{%
-0.156} &
\multicolumn{2}{>{\raggedright\arraybackslash}p{(\columnwidth - 42\tabcolsep) * \real{0.0471} + 2\tabcolsep}@{}}{%
} \\
\end{longtable}
}
\end{landscape}\clearpage

\begin{figure}[H]  
\centering
\includegraphics[width=0.95\linewidth,height=0.82\textheight,keepaspectratio]{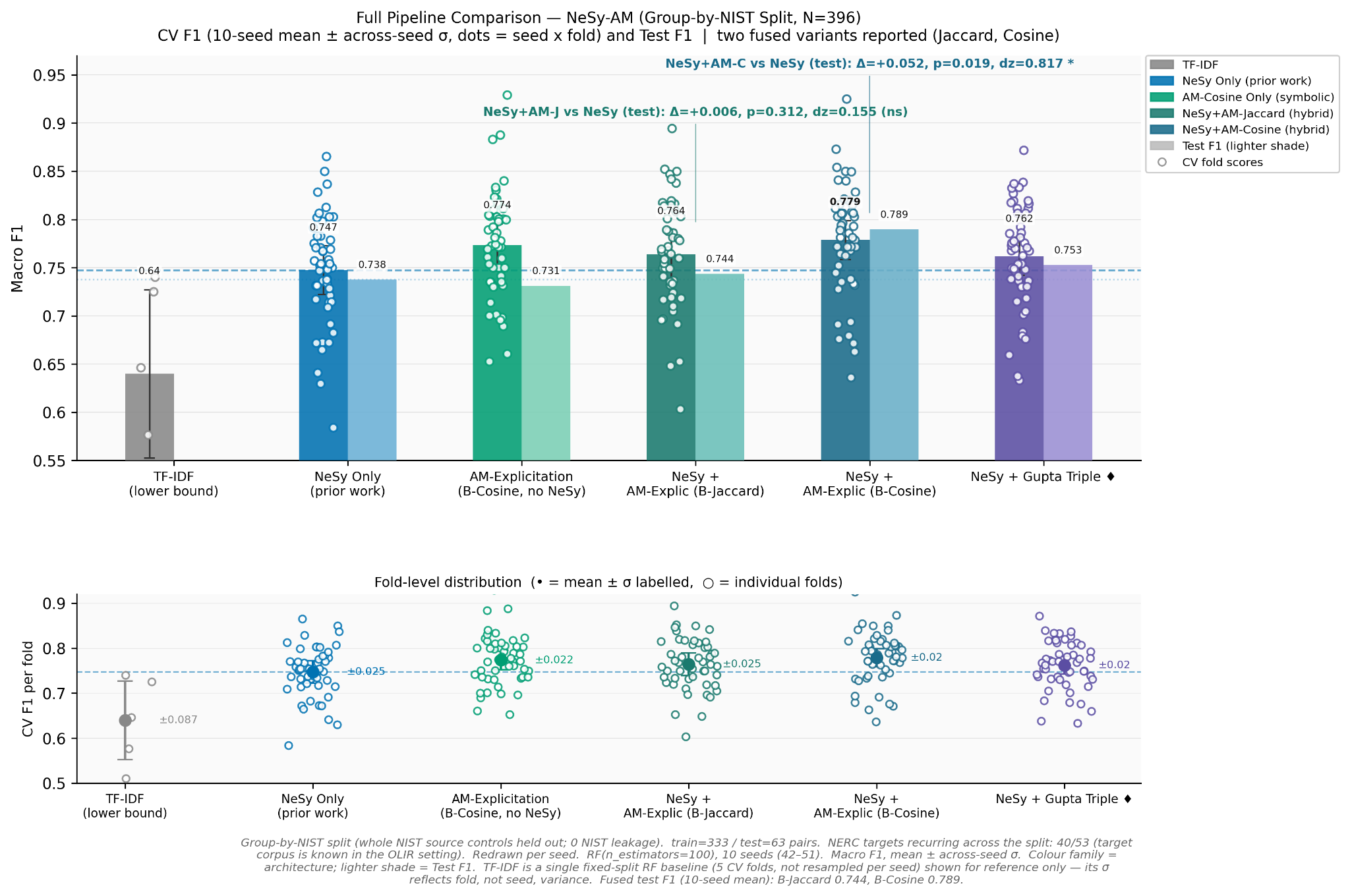}
\caption{NeSy+AM Variants Comparison}
\label{fig:nesy-am-variants}
\end{figure}

\subsection{RQ2 Results -- Does Annotation Methodology
Matter?}\label{rq2-results-does-annotation-methodology-matter}

Annotation methodology matters. Annotating argument structure via
explicitation of enthymemes consistently outperformed all other
methodologies across the repeated cross validation experiments. The
result provides evidence supporting the importance of the explanatory
warrant or `why' in the harmonization task. Extraction does not capture
the `why' because it is never explicit in the source text. However it
can be inferred consistently and its resulting importance is
reproducible across independent experiments.

RQ2 asks, does LLM-based Toulmin enthymeme explicitation produce a
statistically significant increase in classification performance
compared with traditional lexical pattern-based extraction on
cybersecurity standards documents. To investigate RQ2, the annotation
methodologies; extraction (Condition A) and explicitation (Condition B)
were isolated as independent variables, their performance measured, and
the results statistically analyzed, see Tables 7-8 below.

The argumentation models produced by each annotation methodology were
independently run on a complete 10-seed NLP classification pipeline with
both semantic and lexical variants (e.g., B-Jaccard and B-Cosine). The
explicitation annotation methodology outperformed the extraction
methodology with the B-Cosine explicitation variant achieving the
highest performance (CV-F1 0.773, Test-F1 0.731). The complete set of
CV-F1, test F1, and variance results are reported in Table 7 and shown
in Figure 8 below.

\begin{longtable}[]{@{}
  >{\raggedright\arraybackslash}p{(\columnwidth - 10\tabcolsep) * \real{0.1667}}
  >{\raggedright\arraybackslash}p{(\columnwidth - 10\tabcolsep) * \real{0.1667}}
  >{\raggedright\arraybackslash}p{(\columnwidth - 10\tabcolsep) * \real{0.1666}}
  >{\raggedright\arraybackslash}p{(\columnwidth - 10\tabcolsep) * \real{0.1666}}
  >{\raggedright\arraybackslash}p{(\columnwidth - 10\tabcolsep) * \real{0.1666}}
  >{\raggedright\arraybackslash}p{(\columnwidth - 10\tabcolsep) * \real{0.1666}}@{}}
\caption{Annotation F1-CV and F1-Test}\tabularnewline
\toprule\noalign{}
\begin{minipage}[b]{\linewidth}\raggedright
condition
\end{minipage} & \begin{minipage}[b]{\linewidth}\raggedright
cv\_f1\_mean
\end{minipage} & \begin{minipage}[b]{\linewidth}\raggedright
cv\_f1\_std
\end{minipage} & \begin{minipage}[b]{\linewidth}\raggedright
test\_f1\_mean
\end{minipage} & \begin{minipage}[b]{\linewidth}\raggedright
test\_f1\_std
\end{minipage} & \begin{minipage}[b]{\linewidth}\raggedright
n\_seeds
\end{minipage} \\
\midrule\noalign{}
\endfirsthead
\toprule\noalign{}
\begin{minipage}[b]{\linewidth}\raggedright
condition
\end{minipage} & \begin{minipage}[b]{\linewidth}\raggedright
cv\_f1\_mean
\end{minipage} & \begin{minipage}[b]{\linewidth}\raggedright
cv\_f1\_std
\end{minipage} & \begin{minipage}[b]{\linewidth}\raggedright
test\_f1\_mean
\end{minipage} & \begin{minipage}[b]{\linewidth}\raggedright
test\_f1\_std
\end{minipage} & \begin{minipage}[b]{\linewidth}\raggedright
n\_seeds
\end{minipage} \\
\midrule\noalign{}
\endhead
\bottomrule\noalign{}
\endlastfoot
A-Extraction & 0.5307 & 0.0332 & 0.4758 & 0.0826 & 10 \\
B-Jaccard & 0.6658 & 0.0307 & 0.6389 & 0.0541 & 10 \\
B-Cosine & 0.7736 & 0.0218 & 0.7313 & 0.0424 & 10 \\
\end{longtable}

To determine whether explicitation and its variants (Cond. B-Jaccard,
Cond. B-Cosine), significantly outperformed extraction (Cond.
A-Jaccard), a Wilcoxon signed-rank statistical analysis was done on the
results. The statistical analysis shows the results are significant with
explication outperforming extraction, see Table 8 below. Both
explicitation variants (Jaccard and Cosine) significantly outperformed
rule based extraction across CV-F1 and Test F1 (B-Jaccard p=0.001,
B-Cosine p=0.001).

This suggests that the argumentation annotation methodology
(explicitation vs extraction) is the determining factor in the
classification improvements rather than the feature extraction algorithm
(cosine vs Jaccard). Explicitation significantly outperforms extraction.
Table 8 below shows a summary of the RQ2 Wilcoxon signed rank tests.
Figure 6 below summarizes the results by comparing the raw F1
performance scores to the TF-IDF model score added to all figures as a
common baseline consistent across experiments.

\begin{longtable}[]{@{}
  >{\raggedright\arraybackslash}p{(\columnwidth - 12\tabcolsep) * \real{0.1285}}
  >{\raggedright\arraybackslash}p{(\columnwidth - 12\tabcolsep) * \real{0.1160}}
  >{\raggedright\arraybackslash}p{(\columnwidth - 12\tabcolsep) * \real{0.1244}}
  >{\raggedright\arraybackslash}p{(\columnwidth - 12\tabcolsep) * \real{0.1760}}
  >{\raggedright\arraybackslash}p{(\columnwidth - 12\tabcolsep) * \real{0.1963}}
  >{\raggedright\arraybackslash}p{(\columnwidth - 12\tabcolsep) * \real{0.1363}}
  >{\raggedright\arraybackslash}p{(\columnwidth - 12\tabcolsep) * \real{0.1225}}@{}}
\caption{Annotation Type Wilcoxon Significance Test}\tabularnewline
\toprule\noalign{}
\begin{minipage}[b]{\linewidth}\raggedright
condition
\end{minipage} & \begin{minipage}[b]{\linewidth}\raggedright
Metric
\end{minipage} & \begin{minipage}[b]{\linewidth}\raggedright
n\_seeds
\end{minipage} & \begin{minipage}[b]{\linewidth}\raggedright
extraction\_mean
\end{minipage} & \begin{minipage}[b]{\linewidth}\raggedright
explicitation\_mean
\end{minipage} & \begin{minipage}[b]{\linewidth}\raggedright
wilcoxon\_p
\end{minipage} & \begin{minipage}[b]{\linewidth}\raggedright
sig\_flag
\end{minipage} \\
\midrule\noalign{}
\endfirsthead
\toprule\noalign{}
\begin{minipage}[b]{\linewidth}\raggedright
condition
\end{minipage} & \begin{minipage}[b]{\linewidth}\raggedright
Metric
\end{minipage} & \begin{minipage}[b]{\linewidth}\raggedright
n\_seeds
\end{minipage} & \begin{minipage}[b]{\linewidth}\raggedright
extraction\_mean
\end{minipage} & \begin{minipage}[b]{\linewidth}\raggedright
explicitation\_mean
\end{minipage} & \begin{minipage}[b]{\linewidth}\raggedright
wilcoxon\_p
\end{minipage} & \begin{minipage}[b]{\linewidth}\raggedright
sig\_flag
\end{minipage} \\
\midrule\noalign{}
\endhead
\bottomrule\noalign{}
\endlastfoot
B-Jaccard & cv\_f1 & 10 & 0.5307 & 0.6658 & 0.001 & * \\
B-Jaccard & test\_f1 & 10 & 0.4758 & 0.6389 & 0.001 & * \\
B-Cosine & cv\_f1 & 10 & 0.5307 & 0.7736 & 0.001 & * \\
B-Cosine & test\_f1 & 10 & 0.4758 & 0.7313 & 0.001 & * \\
\end{longtable}

\phantomsection\label{_Toc237866650}{}Figure Extraction versus
Explicitation Performance

\begin{figure}[H]  
\centering
\includegraphics[width=0.95\linewidth,height=0.82\textheight,keepaspectratio]{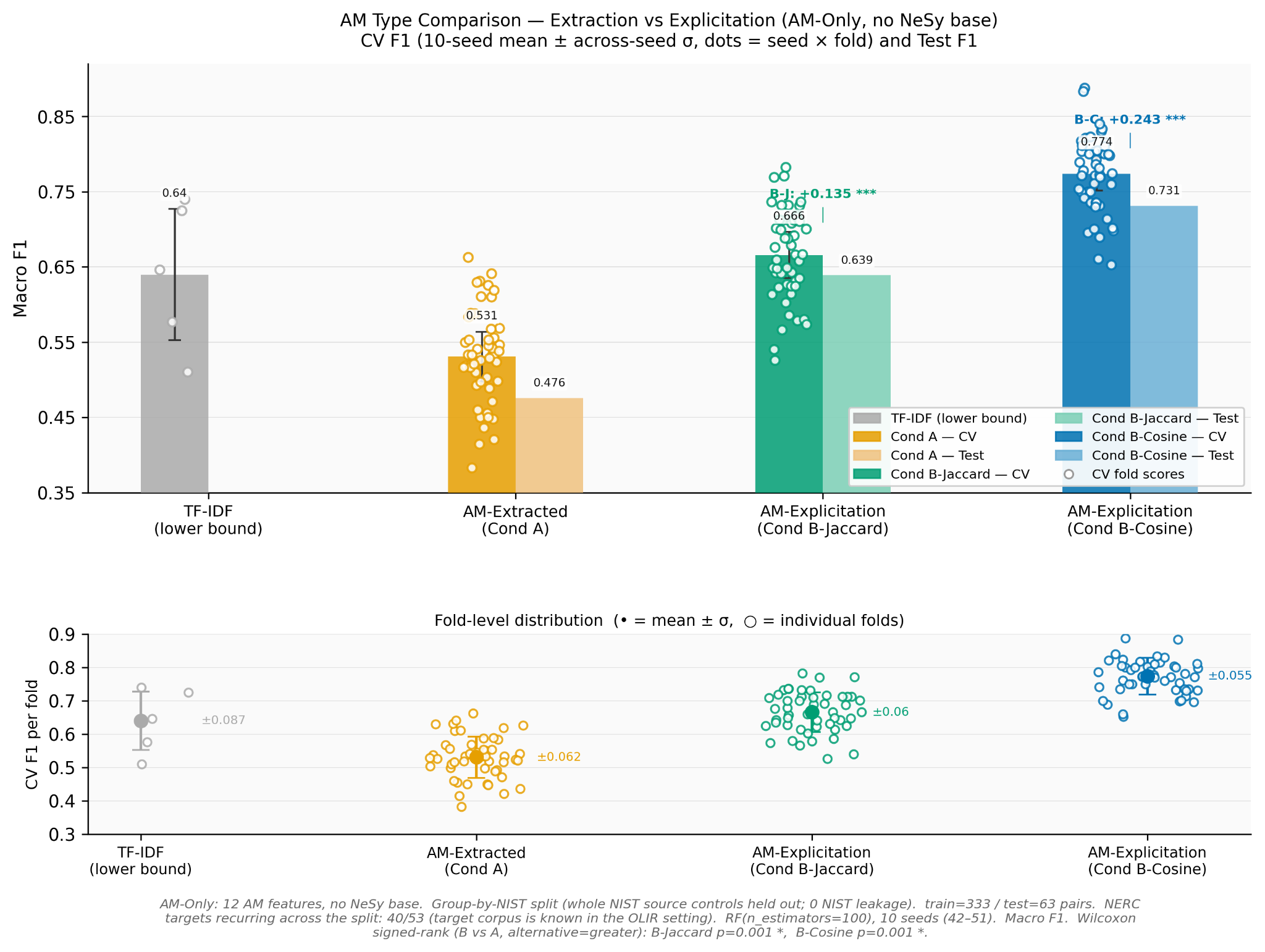}
\caption{Extraction versus Explicitation Performance}
\label{fig:extraction-explicitation}
\end{figure}

\subsection{\texorpdfstring{RQ3 Results -- How Much Argumentation
Structure Is Required?
}{RQ3 Results -- How Much Argumentation Structure Is Required? }}\label{rq3-results-how-much-argumentation-structure-is-required}

A small, consistent group of Toulmin argumentation features provide
sufficient structure to significantly improve harmonization results. RQ3
investigated which AM features added discriminative signal without
confounding the existing NeSy features in the fused model and if these
results were reliable across seeded experiments.

A post-hoc feature importance permutation test was performed to measure
the classification importance of the full 46 feature set, including all
12 AM features of the fused NeSy+AM-cosine model, were measured. Each
feature's directional contribution, magnitude, and CI intervals (95\%)
showing the variability of the mean F1-drop across repeated permutations
was done and is summarized in Figure 7 below and detailed in the
Supplementary Data Appendix.

The permutation ranking highlights both the predictive importance of the
AM features within the fused model and that a small subset of Toulmin
derived AM features (warrants, roles, qualifiers) dominate the models
predictive power as described by Gupta et al. (2024). The top ranked
feature in the fused model is from the AM subset (warrant
similarity=0.033), four of the 12 AM features rank in the top quartile
of permutation importance, and two of the three Toulmin theory based
Gupta-triple features are in the top 50\%, see Supplemental Data
Appendix for raw data and summarized in Figure 7 below.

Additionally, the ranking reliability of the Toulmin canonical roles and
the empirical feature permutation importance across the full 10-seeded
experiments was measured. Kendall's coefficient of concordance (W),
tie-corrected with the standard chi-square significance test was used to
measure the reliability of the rankings. A chi-square approximation was
done to determine if the strength of the reliability (W) across
experiment seeds was statistically significant. The results indicate the
ranking is reliable, with strong concordance agreement that is
statistically significant (W=0.742, Jaccard, p\textless0.001; W=0.458,
Cosine, p=0.0011), see Table 9 and Figure 7-8 below.

\begin{figure}[H]  
\centering
\includegraphics[width=0.95\linewidth,height=0.82\textheight,keepaspectratio]{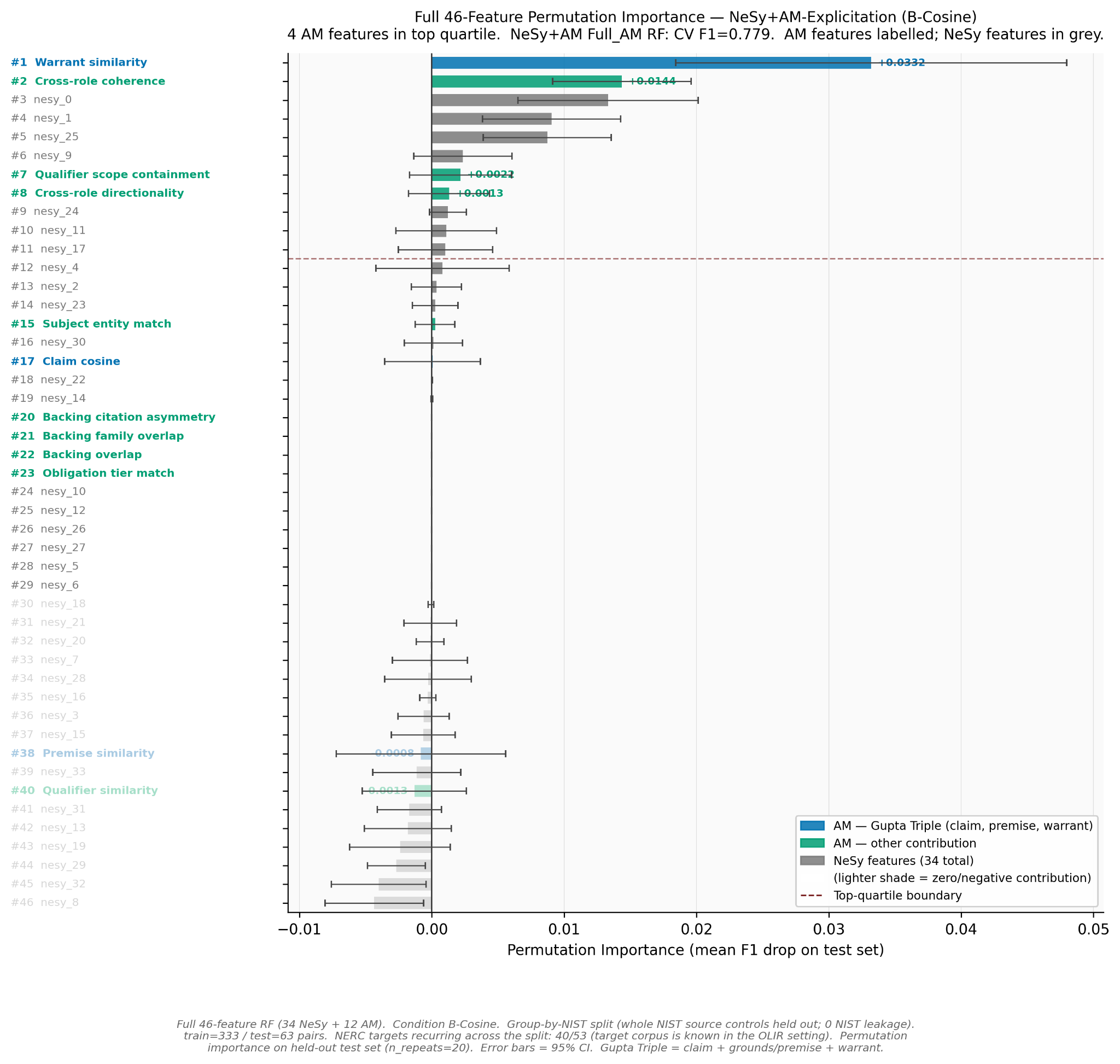}
\caption{Feature Permutation Importance}
\label{fig:feature-permutation-importance}
\end{figure}

\clearpage\begin{landscape}
{\scriptsize\setlength{\tabcolsep}{2pt}\renewcommand{\arraystretch}{0.92}
\begin{longtable}[]{@{}
  >{\raggedright\arraybackslash}p{(\columnwidth - 18\tabcolsep) * \real{0.0842}}
  >{\raggedright\arraybackslash}p{(\columnwidth - 18\tabcolsep) * \real{0.0985}}
  >{\raggedright\arraybackslash}p{(\columnwidth - 18\tabcolsep) * \real{0.0868}}
  >{\raggedright\arraybackslash}p{(\columnwidth - 18\tabcolsep) * \real{0.0668}}
  >{\raggedright\arraybackslash}p{(\columnwidth - 18\tabcolsep) * \real{0.0866}}
  >{\raggedright\arraybackslash}p{(\columnwidth - 18\tabcolsep) * \real{0.0578}}
  >{\raggedright\arraybackslash}p{(\columnwidth - 18\tabcolsep) * \real{0.0963}}
  >{\raggedright\arraybackslash}p{(\columnwidth - 18\tabcolsep) * \real{0.0968}}
  >{\raggedright\arraybackslash}p{(\columnwidth - 18\tabcolsep) * \real{0.2021}}
  >{\raggedright\arraybackslash}p{(\columnwidth - 18\tabcolsep) * \real{0.1241}}@{}}
\caption{AM Theory Alignment}\tabularnewline
\toprule\noalign{}
\begin{minipage}[b]{\linewidth}\raggedright
Cond.
\end{minipage} & \begin{minipage}[b]{\linewidth}\raggedright
W
\end{minipage} & \begin{minipage}[b]{\linewidth}\raggedright
chi2
\end{minipage} & \begin{minipage}[b]{\linewidth}\raggedright
df
\end{minipage} & \begin{minipage}[b]{\linewidth}\raggedright
p
\end{minipage} & \begin{minipage}[b]{\linewidth}\raggedright
Sig.
\end{minipage} & \begin{minipage}[b]{\linewidth}\raggedright
seeds
\end{minipage} & \begin{minipage}[b]{\linewidth}\raggedright
N
\end{minipage} & \begin{minipage}[b]{\linewidth}\raggedright
Modal top role
\end{minipage} & \begin{minipage}[b]{\linewidth}\raggedright
Modal seed count
\end{minipage} \\
\midrule\noalign{}
\endfirsthead
\toprule\noalign{}
\begin{minipage}[b]{\linewidth}\raggedright
Cond.
\end{minipage} & \begin{minipage}[b]{\linewidth}\raggedright
W
\end{minipage} & \begin{minipage}[b]{\linewidth}\raggedright
chi2
\end{minipage} & \begin{minipage}[b]{\linewidth}\raggedright
df
\end{minipage} & \begin{minipage}[b]{\linewidth}\raggedright
p
\end{minipage} & \begin{minipage}[b]{\linewidth}\raggedright
Sig.
\end{minipage} & \begin{minipage}[b]{\linewidth}\raggedright
seeds
\end{minipage} & \begin{minipage}[b]{\linewidth}\raggedright
N
\end{minipage} & \begin{minipage}[b]{\linewidth}\raggedright
Modal top role
\end{minipage} & \begin{minipage}[b]{\linewidth}\raggedright
Modal seed count
\end{minipage} \\
\midrule\noalign{}
\endhead
\bottomrule\noalign{}
\endlastfoot
B-Jaccard & 0.7422 & 29.6884 & 4 & 0.0 & * & 10 & 5 &
am\_warrant\_similarity & 7/10 \\
B-Cosine & 0.458 & 18.32 & 4 & 0.0011 & * & 10 & 5 &
am\_warrant\_similarity & 9/10 \\
\end{longtable}
}
\end{landscape}\clearpage

\begin{figure}[H]  
\centering
\includegraphics[width=0.95\linewidth,height=0.82\textheight,keepaspectratio]{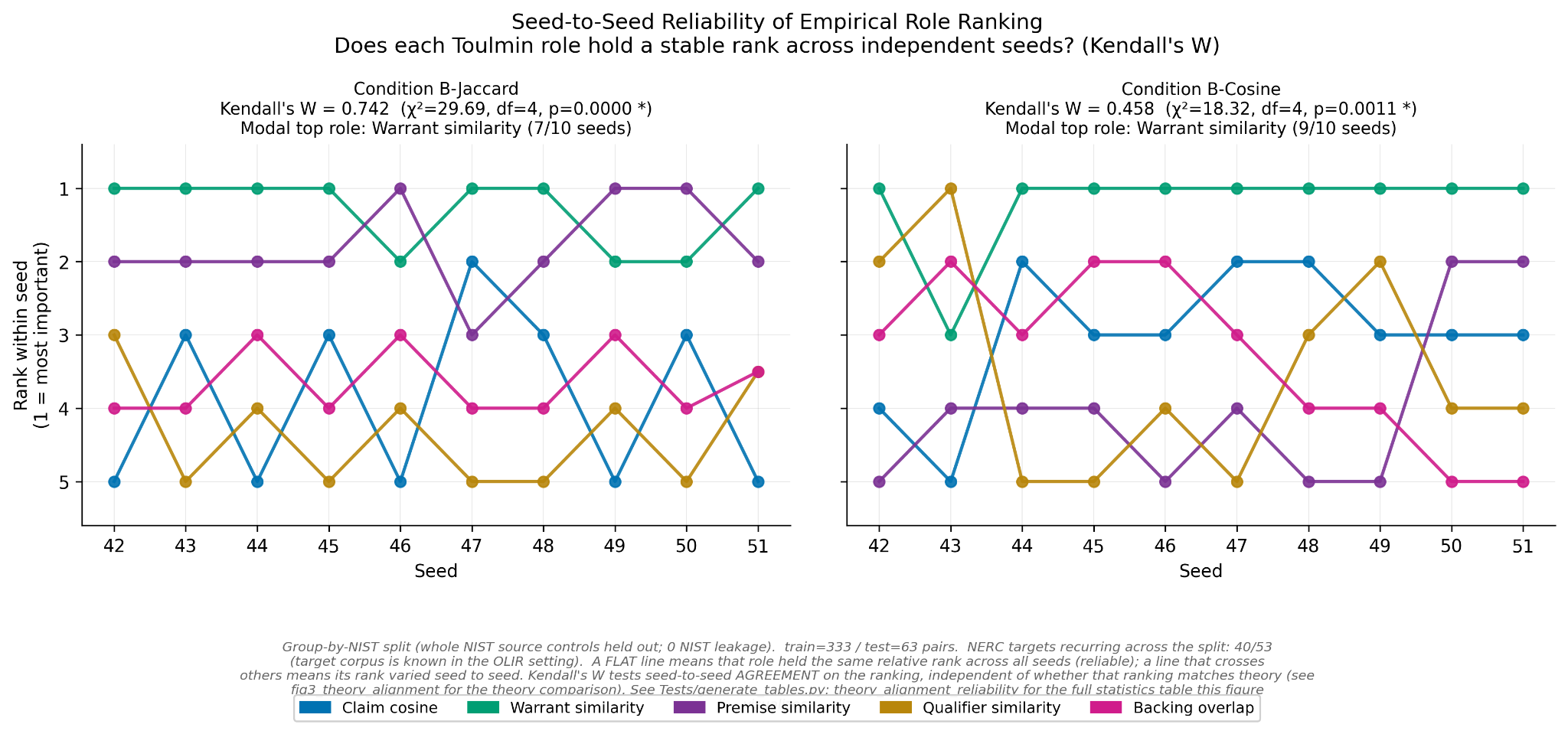}
\caption{Seed-to-Seed Reliability}
\label{fig:seed-reliability}
\end{figure}

\subsection{Validation Results}\label{validation-results}

As described in the Methods section, experimental results were validated
for robustness on a mirror group\_NERC data split. In this group every
test target is unseen with 0 NERC leakage (train=300, test=96, 0 NERC
leakage). These results serve as a robustness validation of the primary
group\_NIST findings in RQ1, see Table 10 below.

The validation replicated cleanly with the results confirming the
classification performance benefit is specific to integrated models
containing the AM features. This is demonstrated by the statistically
significant improvements in the AM+NeSy fused models contrasted with no
AM only ablations reaching statistical significance. The fully fused AM
feature set (NeSy+AM) did not reach statistical significance ($\Delta=$+0.0215,
p=0.065) under group\_NERC validation, only under group\_NIST. However,
the fused NeSy+AM-Gupta fused pipeline did replicate statistical
significance in both groups and more strongly on the group\_NERC
validation ($\Delta=$+0.024, p=0.0186, $d_z$=0.777, 7/10 wins, 95\% CI {[}0.007,
0.043, excluding zero{]}).

Together, the primary experimental results along with the experimental
validation robustness results suggest (a) arguments contain
discriminatory, non-confounding signal, (b) enthymeme explicitation
contributes measurable, significant, classification performance
improvements, and (c) theory based Toulmin argument components
encapsulated in the Gupta-triple (Gupta et al., 2024) are potentially
the more robust source of classification performance in the integrated
NeSy+AM models.

\clearpage\begin{landscape}
{\scriptsize\setlength{\tabcolsep}{2pt}\renewcommand{\arraystretch}{0.92}
\begin{longtable}[]{@{}
  >{\raggedright\arraybackslash}p{(\columnwidth - 32\tabcolsep) * \real{0.0577}}
  >{\raggedright\arraybackslash}p{(\columnwidth - 32\tabcolsep) * \real{0.0919}}
  >{\raggedright\arraybackslash}p{(\columnwidth - 32\tabcolsep) * \real{0.0606}}
  >{\raggedright\arraybackslash}p{(\columnwidth - 32\tabcolsep) * \real{0.0433}}
  >{\raggedright\arraybackslash}p{(\columnwidth - 32\tabcolsep) * \real{0.0511}}
  >{\raggedright\arraybackslash}p{(\columnwidth - 32\tabcolsep) * \real{0.0733}}
  >{\raggedright\arraybackslash}p{(\columnwidth - 32\tabcolsep) * \real{0.0560}}
  >{\raggedright\arraybackslash}p{(\columnwidth - 32\tabcolsep) * \real{0.0619}}
  >{\raggedright\arraybackslash}p{(\columnwidth - 32\tabcolsep) * \real{0.0721}}
  >{\raggedright\arraybackslash}p{(\columnwidth - 32\tabcolsep) * \real{0.0750}}
  >{\raggedright\arraybackslash}p{(\columnwidth - 32\tabcolsep) * \real{0.0395}}
  >{\raggedright\arraybackslash}p{(\columnwidth - 32\tabcolsep) * \real{0.0441}}
  >{\raggedright\arraybackslash}p{(\columnwidth - 32\tabcolsep) * \real{0.0346}}
  >{\raggedright\arraybackslash}p{(\columnwidth - 32\tabcolsep) * \real{0.0597}}
  >{\raggedright\arraybackslash}p{(\columnwidth - 32\tabcolsep) * \real{0.0590}}
  >{\raggedright\arraybackslash}p{(\columnwidth - 32\tabcolsep) * \real{0.0717}}
  >{\raggedright\arraybackslash}p{(\columnwidth - 32\tabcolsep) * \real{0.0486}}@{}}
\caption{Split-strategy Validation F1-CV and F1-test}\tabularnewline
\toprule\noalign{}
\begin{minipage}[b]{\linewidth}\raggedright
Cond.
\end{minipage} & \begin{minipage}[b]{\linewidth}\raggedright
Config.
\end{minipage} & \begin{minipage}[b]{\linewidth}\raggedright
Basel.
\end{minipage} & \begin{minipage}[b]{\linewidth}\raggedright
metric
\end{minipage} & \begin{minipage}[b]{\linewidth}\raggedright
n\_seeds
\end{minipage} & \begin{minipage}[b]{\linewidth}\raggedright
baseline\_mean
\end{minipage} & \begin{minipage}[b]{\linewidth}\raggedright
am\_mean
\end{minipage} & \begin{minipage}[b]{\linewidth}\raggedright
mean\_delta
\end{minipage} & \begin{minipage}[b]{\linewidth}\raggedright
delta\_ci95\_low
\end{minipage} & \begin{minipage}[b]{\linewidth}\raggedright
delta\_ci95\_high
\end{minipage} & \begin{minipage}[b]{\linewidth}\raggedright
Wins
\end{minipage} & \begin{minipage}[b]{\linewidth}\raggedright
losses
\end{minipage} & \begin{minipage}[b]{\linewidth}\raggedright
ties
\end{minipage} & \begin{minipage}[b]{\linewidth}\raggedright
wilcoxon\_p
\end{minipage} & \begin{minipage}[b]{\linewidth}\raggedright
cohens\_dz
\end{minipage} & \begin{minipage}[b]{\linewidth}\raggedright
rank\_biserial\_r
\end{minipage} & \begin{minipage}[b]{\linewidth}\raggedright
sig\_flag
\end{minipage} \\
\midrule\noalign{}
\endfirsthead
\toprule\noalign{}
\begin{minipage}[b]{\linewidth}\raggedright
Cond.
\end{minipage} & \begin{minipage}[b]{\linewidth}\raggedright
Config.
\end{minipage} & \begin{minipage}[b]{\linewidth}\raggedright
Basel.
\end{minipage} & \begin{minipage}[b]{\linewidth}\raggedright
metric
\end{minipage} & \begin{minipage}[b]{\linewidth}\raggedright
n\_seeds
\end{minipage} & \begin{minipage}[b]{\linewidth}\raggedright
baseline\_mean
\end{minipage} & \begin{minipage}[b]{\linewidth}\raggedright
am\_mean
\end{minipage} & \begin{minipage}[b]{\linewidth}\raggedright
mean\_delta
\end{minipage} & \begin{minipage}[b]{\linewidth}\raggedright
delta\_ci95\_low
\end{minipage} & \begin{minipage}[b]{\linewidth}\raggedright
delta\_ci95\_high
\end{minipage} & \begin{minipage}[b]{\linewidth}\raggedright
Wins
\end{minipage} & \begin{minipage}[b]{\linewidth}\raggedright
losses
\end{minipage} & \begin{minipage}[b]{\linewidth}\raggedright
ties
\end{minipage} & \begin{minipage}[b]{\linewidth}\raggedright
wilcoxon\_p
\end{minipage} & \begin{minipage}[b]{\linewidth}\raggedright
cohens\_dz
\end{minipage} & \begin{minipage}[b]{\linewidth}\raggedright
rank\_biserial\_r
\end{minipage} & \begin{minipage}[b]{\linewidth}\raggedright
sig\_flag
\end{minipage} \\
\midrule\noalign{}
\endhead
\bottomrule\noalign{}
\endlastfoot
B-Jaccard & NeSy + Full AM (12 features) & NeSy\_Only & cv & 10 & 0.7431
& 0.7717 & 0.0286 & 0.0242 & 0.0323 & 10 & 0 & 0 & 0.001 & 4.206 & 1.0 &
* \\
B-Jaccard & NeSy + Full AM (12 features) & NeSy\_Only & test & 10 &
0.7436 & 0.7484 & 0.0048 & -0.0187 & 0.0273 & 6 & 3 & 1 & 0.3262 & 0.123
& 0.2 & \\
B-Jaccard & NeSy + Gupta Triple (claim/premise/warrant) & NeSy\_Only &
cv & 10 & 0.7431 & 0.7655 & 0.0224 & 0.0134 & 0.0319 & 10 & 0 & 0 &
0.001 & 1.414 & 1.0 & * \\
B-Jaccard & NeSy + Gupta Triple (claim/premise/warrant) & NeSy\_Only &
test & 10 & 0.7436 & 0.7613 & 0.0177 & -0.0007 & 0.0377 & 6 & 4 & 0 &
0.0967 & 0.541 & 0.491 & \\
B-Jaccard & AM Only (no NeSy base) & NeSy\_Only & cv & 10 & 0.7431 &
0.6588 & -0.0843 & -0.1001 & -0.0667 & 0 & 10 & 0 & 1.0 & -2.917 & -1.0
& \\
B-Jaccard & AM Only (no NeSy base) & NeSy\_Only & test & 10 & 0.7436 &
0.6506 & -0.093 & -0.1499 & -0.0414 & 2 & 8 & 0 & 0.9971 & -1.01 &
-0.891 & \\
B-Cosine & NeSy + Full AM (12 features) & NeSy\_Only & cv & 10 & 0.7431
& 0.7759 & 0.0328 & 0.0262 & 0.0383 & 10 & 0 & 0 & 0.001 & 3.225 & 1.0 &
* \\
B-Cosine & NeSy + Full AM (12 features) & NeSy\_Only & test & 10 &
0.7436 & 0.7651 & 0.0215 & -0.0043 & 0.0478 & 6 & 4 & 0 & 0.0654 & 0.488
& 0.564 & \\
B-Cosine & NeSy + Gupta Triple (claim/premise/warrant) & NeSy\_Only & cv
& 10 & 0.7431 & 0.7745 & 0.0314 & 0.0215 & 0.0415 & 10 & 0 & 0 & 0.001 &
1.83 & 1.0 & * \\
B-Cosine & NeSy + Gupta Triple (claim/premise/warrant) & NeSy\_Only &
test & 10 & 0.7436 & 0.7674 & 0.0238 & 0.0067 & 0.0428 & 7 & 3 & 0 &
0.0186 & 0.777 & 0.745 & * \\
B-Cosine & AM Only (no NeSy base) & NeSy\_Only & cv & 10 & 0.7431 &
0.7644 & 0.0213 & 0.0074 & 0.0358 & 9 & 1 & 0 & 0.0137 & 0.875 & 0.782 &
* \\
B-Cosine & AM Only (no NeSy base) & NeSy\_Only & test & 10 & 0.7436 &
0.7545 & 0.0109 & -0.0201 & 0.0423 & 6 & 4 & 0 & 0.3125 & 0.207 & 0.2
& \\
\end{longtable}
}
\end{landscape}\clearpage

\section{Discussion}\label{discussion}

\subsection{Argument Structures are Complementary Semantic
Representations}\label{argument-structures-are-complementary-semantic-representations}

The central finding of this research is that argument derived
representations can improve classification performance when combined
with neural representations. This result supports a view of specialized
normative text in which meaning is not exhausted by the entities,
actions and topics mentioned in a requirement. Normative requirements
also encode relationships between conditions, evidence, reasoning, and
obligations

Two requirements may therefore be semantically related at the lexical or
topical level while differing in the relationship between their
supporting grounds and normative conclusions. Conversely, requirements
may express similar reasoning using different vocabulary. Argument
structure provides a representation through which these relationships
can be made explicit.

The results do not establish that Toulmin representations are
universally superior to semantic embeddings, rather they suggest that
the two representations encode partially complementary information and
that argument structures are complementary semantic representations.

\subsection{`Why' Warrants May Matter}\label{why-warrants-may-matter}

The strongest feature selection results involve warrant related
representations. A warrant connects grounds to a claim and provides the
mechanism to explain `why' they are related. In normative text, this
relationship can be particularly informative because the same action may
be required for different reasons, while different actions may be
required under similar conditions. For example, two requirements may
both concern privileged access but differ in the reasoning that connects
privileged access to the prescribed control. Conversely, two
requirements may prescribe different surface actions while relying on a
similar underlying justification.

A conventional vector embedding may encode the semantic relatedness of
the tests without explicitly representing this distinction. The warrant
representation provides one mechanism for making that relationship
computationally explicit. This interpretation remains a hypothesis
rather than a demonstrated causal explanation of the model's behavior.
Further work using human annotated argument structures and broader
domains is necessary to determine whether warrant information
consistently provides this advantage.

\subsection{Argument Structure versus Domain Specific
Features}\label{argument-structure-versus-domain-specific-features}

An important question is whether the observed gains arise from
argumentation or simply from additional engineered features. The feature
analysis suggests that warrant and grounds related features provide
substantial predictive signal. However, several features in the full
representation are specific to the cybersecurity standards domain,
including obligation tiers and subject entity relationships. Therefore,
the present experiments cannot be interpreted as completely isolating
the contributions of Toulmin argumentation.

However, there is direct initial evidence that argumentative structure
rather than additional domain specific features is the causal agent. The
Gupta-triple (claim, premise, warrant) matches or beats the full AM
domain specific feature set on test F1, on both splits. If the gains
were being driven by the domain specific additions, the full AM feature
set would be expected to outperform the argument theory based subset.
The empirical results show that this is not the case and provide
initial, but incomplete evidence that argumentation theory provides the
performance improvements.

\section{Limitations}\label{limitations}

\subsection{Argument Representation Is Not Independently Gold
Standard
Annotated}\label{argument-representation-is-not-independently-gold-standard-annotated}

LLM-generated Toulmin representations are not evaluated against an
independently constructed human annotated Toulmin corpus. An LLM
generated confidence level is used and accordingly, the study does not
establish the accuracy of the argument extraction or explication process
itself. Therefore, the appropriate interpretation should be that the
generated argument derived representation provides useful information
for a downstream classification task. A future study should collect a
human annotated sample evaluated by multiple expert annotators and
measure inter-annotator agreement along with component level extraction
accuracy using Cohen's kappa, Fleiss' kappa or a similar metric.

\subsection{Dataset Size and Domain}\label{dataset-size-and-domain}

The benchmark contains 396 examples and is limited to cybersecurity
standards, Although the domain provides a useful and relevant testbed
for semantic alignment, the results cannot establish that the same
representations will produce comparable improvements in other domains.
Future evaluation should include additional normative domains such as
regulations, legal requirements, technical standards, or organizational
policies.

\subsection{Statistical Inference}\label{statistical-inference}

Repeated seeded runs provide information about model variability but
seeds are not equivalent to independent datasets. Seed-level
significance establishes internal robustness (no dependence on any
particular split draw) but does not establish external validity on if
the results would hold in another domain. However, while not a different
dataset, the split is redrawn per seed and the mirror-split design does
validate the results using a differently structured draw for a stronger
check than seeds within one split alone.

\section{\texorpdfstring{Conclusion and Future Work
}{Conclusion and Future Work }}\label{conclusion-and-future-work}

Regulatory documents use specific formats (e.g., NIST-CSF, NERC-CIP) to
encode justificatory and inferential reasoning within their specific
domains. These reasoning types are also the essential properties of
formal arguments. Framing regulatory documents as generalized arguments
offers a bridge between regulatory document formats and the fields of
computational argumentation and NLP. Using cybersecurity standards as an
empirical testbed, we combined neural semantic representations with
Toulmin-derived argument features in neuro-symbolic harmonization
pipeline to investigate whether explicit argument structure can provide
complementary information for semantic alignment of specialized
normative texts.

We bridge these concepts by implementing a novel NeSy+AM Toulmin
argumentation framework to empirically address three research questions.
The reusable NLP-AM framework, see Figures 3-4, and the experimental
results from each research question provide the core contributions this
research makes: (a) significant empirical validation of the fused
NeSy+AM-cosine model replicated across independent dataset splits, see
Figure 5, (b) significant empirical validation of enthymeme
reconstruction annotation methodology, see Figure 6, and (c) a
replicable, statistically significant group of Toulmin theory
argumentation features provides sufficient signal to improve
harmonization performance, see Figure 7-8.

These results indicate that argument structure is a complementary and
effective representation for modeling requirements as arguments which
captures signal that statistically improves classification performance.
Further, this research provides insights into the mechanisms driving
these improvements through post-hoc feature permutation rankings and
testing the significance of enthymeme reconstruction. We find a compact
explicit argumentation structure may capture substantial harmonization
alignment relevant information. Taken together, these findings provide
preliminary evidence that argumentative structure can serve as an
intermediate representation for harmonization alignment rather than
solely as an extraction target. The study does not establish the
accuracy of LLM generated argument annotations or demonstrate domain
independent generalization. Human annotated argument data and larger
more diverse benchmarks are needed to address these questions.

While these findings apply directly to automating standards
harmonization and improving classification accuracy to improve the
quality, efficiency and economics of compliance, an important industry
application and an essential requirement for critical infrastructure -
the broader implication is that semantic alignment may benefit from
representations that capture not only what specialized texts discuss,
but also how claims are supported and related. Argument-aware
representations therefore offer a promising direction for future
research in semantic alignment, argument mining, retrieval, natural
language inference, and reasoning over normative text.

\section*{Acknowledgements}\label{acknowledgements}

The author is deeply grateful to \textbf{Dr. Alejandra J. Magana, W.C.
Furnas Professor in Enterprise Excellence, Professor of Applied and
Creative Computing at Purdue University}, for her careful review of
earlier versions of this manuscript and for her insightful comments,
scholarly guidance, and editorial recommendations. The author also
gratefully acknowledges \textbf{Brice Williams, Cybersecurity Practice
Lead and Senior Security Architect at SysLogic, Inc.,} for his extensive
expertise in cybersecurity and information security, as well as for his
valuable industry perspective and thoughtful guidance throughout the
development of this work. Their expertise, perspectives, and
contributions were instrumental in strengthening the rigor, clarity, and
overall quality of this work.

\section*{References}\label{references}

Abdeen, B., Al-Shaer, E., Singhal, A., Khan, L., \& Hamlen, K. (2023).
SMET: Semantic mapping of CVE to ATT\&CK and its application to
cybersecurity. Springer.

Altmann, S. A., Tolo\\c{s}i, L., Sander, O., \& Lengauer, T. (2010).
Permutation importance: A corrected feature importance measure.
Bioinformatics, 26(10), 1340--1347. \href{http://doi.org}{\ul{doi.org}}

Agarwal, V., Butler, C., Degenaro, L., Kumar, A., Sailer, A., \&
Steinder, G. (2022). Compliance-as-code for cybersecurity automation in
hybrid cloud. 2022 IEEE 15th International Conference on Cloud Computing
(CLOUD). https://doi.org/10.1109/CLOUD55607.2022

Gupta, A., Zuckerman, E., \& O\textquotesingle Connor, B. (2024).
\emph{Harnessing Toulmin\textquotesingle s theory for zero-shot argument
explication}. In L.-W. Ku, A. Martins, \& V. Srikumar (Eds.),
\emph{Proceedings of the 62nd Annual Meeting of the Association for
Computational Linguistics (Volume 1: Long Papers)} (pp. 10259--10276).
Association for Computational Linguistics.
https://doi.org/10.18653/v1/2024.acl-long.552

Haley, C., Laney, R., Moffett, J., \& Nuseibeh, B. (2008).
\emph{Security requirements engineering: A framework for representation
and analysis}. \emph{IEEE Transactions on Software Engineering, 34}(1),
133--153. https://doi.org/10.1109/TSE.2007.70754

Hollander, M., Wolfe, D. A., \& Chicken, E. (2014). Nonparametric
statistical methods (3rd ed.). Wiley.

Lawrence, J., \& Reed, C. (2019). Argument mining: A survey.
Computational Linguistics, 45(4), 765--818.
https://doi.org/10.1162/coli\_a\_00364

Lytos, A., Lagkas, T., Sarigiannidis, P. G., \& Bontcheva, K. (2019).
The evolution of argumentation mining: From models to social media and
emerging tools. Information Processing \& Management, 56(6), Article
102055.
\href{https://doi.org/10.1016/j.ipm.2019.102055}{\ul{https://doi.org/10.1016/j.ipm.2019.102055}}

Martins, B. F., Serrano Gil, L. J., Reyes Rom\\'{a}n, J. F., Panach, J. I.,
Pastor, O., Hadad, M., \& Rochwerger, B. (2022). A Framework for
conceptual characterization of ontologies and its application in the
cybersecurity domain. Software and Systems Modeling.
https://doi.org/10.1007/s10270-022-01013-0

National Institute of Standards and Technology. (2023). Informative
Reference Catalog: Reference ID 90. Retrieved February 8, 2025, from
https://csrc.nist.gov/projects/olir/informative-reference-catalog/details?referenceId=90\#

Olifer, D., Goranin, N., Cenys, A., Kaceniauskas, A., \& Janulevicius,
J. (2019). Defining the minimum security baseline in a multiple security
standards environment by graph theory techniques. Applied Sciences,
9(4), 748. https://doi.org/10.3390/app9040748

Peldszus, A., \& Stede, M. (2013). From argument diagrams to
argumentation mining in texts: A survey. International Journal of
Cognitive Informatics and Natural Intelligence (IJCINI), 7(1), 1--31.
https://doi.org/10.4018/jcini.2013010101

Schneier, B. (2018). \emph{Click here to kill everybody: Security and
survival in a hyper-connected world}.
\href{https://nerd.wwnorton.com/ebooks/epub/mlaupdate2021/EPUB/content/2.5.3-chapter22-mla-supplement.xhtml}{\ul{W.
W. Norton \& Company}}.
{[}\href{https://www.cliffsnotes.com/cliffs-questions/4725911}{\ul{1}}{]}

Schroeder, W. (2025). A neuro-symbolic semantic analysis approach to the
harmonization of cybersecurity standards in the energy domain
{[}Doctoral dissertation, Purdue University{]}. Purdue University
Graduate School. https://doi.org/10.25394/PGS.30781292

Schroeder, W. N. (2025, February 5). Rosetta: Computational Scripts
(Version 2.0) {[}Source code{]}. GitHub.
{[}https://github.com/bshredder/Rosetta{]}

Viger, T., Diemert, S., \& Foster, O. (2023). \emph{Patterns for
integrating NIST 800-53 controls into security assurance cases}. In
\emph{Computer Safety, Reliability, and Security -- SAFECOMP 2023
Workshops} (pp. 165--175). Springer.

Virtanen, P., Gommers, R., Oliphant, T. E., Haberland, M., Reddy, T.,
Cournapeau, D., Burovski, E., Peterson, P., Weckesser, W., Bright, J.,
van der Walt, S. J., Brett, M., Wilson, J., Millman, K. J., Mayorov, N.,
Nelson, A. R. J., Jones, E., Kern, R., Larson, E., \ldots{} Van
Mulbregt, P. (2020). SciPy 1.0: Fundamental algorithms for scientific
computing in Python. doi.org

\subsection{\texorpdfstring{Appendix A - Explicitation
}{Appendix A - Explicitation }}\label{appendix-a---explicitation}

\subsection{Annotation Prompt}\label{annotation-prompt}

ANNOTATION\_PROMPT = """You are an expert in regulatory compliance text
analysis.

Decompose the following cybersecurity requirement into Toulmin argument
components.

Return ONLY valid JSON -\/- no preamble, no markdown, no explanation.

CRITICAL RULES:

1. Prioritize direct extraction from the text.

2. Use minimal, high-confidence inference ONLY when a component is not
explicitly

stated but is clearly implied by the requirement\textquotesingle s
wording and standard

cybersecurity context.

3. Do not over-infer. Warrants must remain conservative and logical.

4. Return {[}{]} when no reasonable basis (explicit or implicitly clear)
exists.

JSON schema:

\{\{

"claim": "\textless core compliance obligation\textgreater",

"premises": {[}"\textless specific technical condition or
grounds\textgreater"{]},

"warrants": {[}"\textless logical bridge connecting premises to
claim\textgreater"{]},

"qualifiers": {[}"\textless scope marker\textgreater"{]},

"backing": {[}"\textless explicitly named standard\textgreater"{]}

\}\}

Role definitions (inference as fallback):

claim: The primary compliance obligation -\/- what must be done.

Direct extraction or minimal rephrasing from text.

premises: The specific technical condition, fact, or grounds that make
this

control necessary. Can be explicit or clearly implied by the

requirement\textquotesingle s domain context. Must be distinct from the
claim -\/-

not a restatement of it.

warrants: The logical bridge or underlying assumption that connects the

premises (grounds) to the claim. It answers: "Why does this premise

support this claim?" or "What general rule allows us to go from

this premise to this claim?" Keep warrants concise and focused on

the logical connection rather than restating the threat.

DO NOT produce: "Failure to comply may result in security incidents"

or other generic consequence statements.

qualifiers: Explicit or obviously implied scope limitations (who, what,
when).

backing: Standards, regulations, or frameworks explicitly named in the
text

only. No prior knowledge.

Examples:

Input: "The responsible entity shall implement a patch management
program for

applicable Cyber Assets within the Electronic Security Perimeter."

Output:

\{\{

"claim": "The responsible entity shall implement a patch management
program for applicable Cyber Assets within the Electronic Security
Perimeter",

"premises": {[}"Cyber Assets within the Electronic Security Perimeter
are subject to evolving software vulnerabilities"{]},

"warrants": {[}"Timely patching of known vulnerabilities is necessary to
prevent exploitation of Cyber Assets"{]},

"qualifiers": {[}"applicable Cyber Assets", "within the Electronic
Security Perimeter"{]},

"backing": {[}{]}

\}\}

Input: "Senior executives understand their roles and responsibilities."

Output:

\{\{

"claim": "Senior executives understand their roles and
responsibilities",

"premises": {[}"Senior executives have defined roles in the
organization\textquotesingle s cybersecurity program"{]},

"warrants": {[}"Understanding one\textquotesingle s roles and
responsibilities is a prerequisite for effective accountability and
decision-making in cybersecurity governance"{]},

"qualifiers": {[}"Senior executives"{]},

"backing": {[}{]}

\}\}

Input: "Access permissions and authorizations are managed, incorporating
the

principles of least privilege and separation of duties."

Output:

\{\{

"claim": "Access permissions and authorizations are managed
incorporating least privilege and separation of duties",

"premises": {[}"Unrestricted access permissions increase the risk of
unauthorized actions and privilege abuse"{]},

"warrants": {[}"Least privilege and separation of duties limit the blast
radius of compromised credentials or insider threats"{]},

"qualifiers": {[}{]},

"backing": {[}{]}

\}\}

Now decompose this requirement:

Input: "\{text\}"

Output"""

\section{Appendix B -- Declaration of Generative
AI}\label{appendix-b-declaration-of-generative-ai}

During the preparation of this work, the authors used ChatGPT (GPT-4o,
OpenAI) and GitHub Copilot (GPT-5-mini) in order to generate or debug
experimental code, assist with literature identification, and convert
to LaTeX document markup format. After using all tools, the authors 
reviewed and edited the content as needed and take full responsibility 
for the content of the published article.

\section{Appendix C - Supplementary
Data}\label{appendix-c---supplementary-data}

\subsection{\texorpdfstring{Feature Ranking
}{Feature Ranking }}\label{feature-ranking}

\begin{longtable}[]{@{}
  >{\raggedright\arraybackslash}p{(\columnwidth - 10\tabcolsep) * \real{0.0683}}
  >{\raggedright\arraybackslash}p{(\columnwidth - 10\tabcolsep) * \real{0.3467}}
  >{\raggedright\arraybackslash}p{(\columnwidth - 10\tabcolsep) * \real{0.1671}}
  >{\raggedright\arraybackslash}p{(\columnwidth - 10\tabcolsep) * \real{0.2062}}
  >{\raggedright\arraybackslash}p{(\columnwidth - 10\tabcolsep) * \real{0.1029}}
  >{\raggedright\arraybackslash}p{(\columnwidth - 10\tabcolsep) * \real{0.1088}}@{}}
\caption{Supplementary Feature Rank Data}\tabularnewline
\toprule\noalign{}
\begin{minipage}[b]{\linewidth}\raggedright
\textbf{rank}
\end{minipage} & \begin{minipage}[b]{\linewidth}\raggedright
\textbf{feature}
\end{minipage} & \begin{minipage}[b]{\linewidth}\raggedright
\textbf{is\_am\_feature}
\end{minipage} & \begin{minipage}[b]{\linewidth}\raggedright
\textbf{mean\_importance}
\end{minipage} & \begin{minipage}[b]{\linewidth}\raggedright
\textbf{std}
\end{minipage} & \begin{minipage}[b]{\linewidth}\raggedright
\textbf{n\_seeds}
\end{minipage} \\
\midrule\noalign{}
\endfirsthead
\toprule\noalign{}
\begin{minipage}[b]{\linewidth}\raggedright
\textbf{rank}
\end{minipage} & \begin{minipage}[b]{\linewidth}\raggedright
\textbf{feature}
\end{minipage} & \begin{minipage}[b]{\linewidth}\raggedright
\textbf{is\_am\_feature}
\end{minipage} & \begin{minipage}[b]{\linewidth}\raggedright
\textbf{mean\_importance}
\end{minipage} & \begin{minipage}[b]{\linewidth}\raggedright
\textbf{std}
\end{minipage} & \begin{minipage}[b]{\linewidth}\raggedright
\textbf{n\_seeds}
\end{minipage} \\
\midrule\noalign{}
\endhead
\bottomrule\noalign{}
\endlastfoot
1 & am\_warrant\_similarity & True & 0.0332 & 0.02382 & 10 \\
2 & am\_grounds\_warrant\_coherence & True & 0.01436 & 0.00843 & 10 \\
3 & nesy\_0 & False & 0.01332 & 0.011 & 10 \\
4 & nesy\_1 & False & 0.00904 & 0.00843 & 10 \\
5 & nesy\_25 & False & 0.00873 & 0.00779 & 10 \\
6 & nesy\_9 & False & 0.00235 & 0.00598 & 10 \\
7 & am\_qualifier\_scope\_containment & True & 0.00218 & 0.00623 & 10 \\
8 & am\_gwc\_directionality & True & 0.00131 & 0.00495 & 10 \\
9 & nesy\_24 & False & 0.00122 & 0.00223 & 10 \\
10 & nesy\_11 & False & 0.0011 & 0.00614 & 10 \\
11 & nesy\_17 & False & 0.00104 & 0.00573 & 10 \\
12 & nesy\_4 & False & 0.00082 & 0.00813 & 10 \\
13 & nesy\_2 & False & 0.00035 & 0.00305 & 10 \\
14 & nesy\_23 & False & 0.00025 & 0.00279 & 10 \\
15 & am\_subject\_entity\_match & True & 0.00025 & 0.00241 & 10 \\
16 & nesy\_30 & False & 0.00012 & 0.00355 & 10 \\
17 & am\_claim\_cosine & True & 6e-05 & 0.00584 & 10 \\
18 & nesy\_22 & False & 2e-05 & 6e-05 & 10 \\
19 & nesy\_14 & False & 0.0 & 0.00012 & 10 \\
20 & am\_backing\_citation\_asymmetry & True & 0.0 & 0.0 & 10 \\
21 & am\_backing\_family\_overlap & True & 0.0 & 0.0 & 10 \\
22 & am\_backing\_overlap & True & 0.0 & 0.0 & 10 \\
23 & am\_obligation\_tier\_match & True & 0.0 & 0.0 & 10 \\
24 & nesy\_10 & False & 0.0 & 0.0 & 10 \\
25 & nesy\_12 & False & 0.0 & 0.0 & 10 \\
26 & nesy\_26 & False & 0.0 & 0.0 & 10 \\
27 & nesy\_27 & False & 0.0 & 0.0 & 10 \\
28 & nesy\_5 & False & 0.0 & 0.0 & 10 \\
29 & nesy\_6 & False & 0.0 & 0.0 & 10 \\
30 & nesy\_18 & False & -6e-05 & 0.00034 & 10 \\
31 & nesy\_21 & False & -0.00011 & 0.0032 & 10 \\
32 & nesy\_20 & False & -0.00012 & 0.00169 & 10 \\
33 & nesy\_7 & False & -0.00014 & 0.00458 & 10 \\
34 & nesy\_28 & False & -0.00028 & 0.0053 & 10 \\
35 & nesy\_16 & False & -0.0003 & 0.00099 & 10 \\
36 & nesy\_3 & False & -0.00061 & 0.00311 & 10 \\
37 & nesy\_15 & False & -0.00065 & 0.0039 & 10 \\
38 & am\_premise\_similarity & True & -0.00082 & 0.01034 & 10 \\
39 & nesy\_33 & False & -0.00114 & 0.00535 & 10 \\
40 & am\_qualifier\_similarity & True & -0.00131 & 0.00634 & 10 \\
41 & nesy\_31 & False & -0.00169 & 0.00392 & 10 \\
42 & nesy\_13 & False & -0.00181 & 0.00531 & 10 \\
43 & nesy\_19 & False & -0.0024 & 0.00614 & 10 \\
44 & nesy\_29 & False & -0.00267 & 0.00352 & 10 \\
45 & nesy\_32 & False & -0.00401 & 0.00577 & 10 \\
46 & nesy\_8 & False & -0.00435 & 0.00601 & 10 \\
\end{longtable}

\section{Appendix D - Formulas}\label{appendix-d---formulas}

The equations below continue the paper-wide equation numbering (4-18).

\setcounter{equation}{3}

\begin{equation}
D_{\mathrm{Euclidean}}(A,B)
= \sqrt{\sum_{i=1}^{n}\left(A_i-B_i\right)^2}
\label{eq:euclidean}
\end{equation}

\begin{equation}
\mathrm{CosineSimilarity}(\cos\theta)
= \frac{A\cdot B}{\lVert A\rVert\,\lVert B\rVert}
\label{eq:cosine}
\end{equation}

\begin{equation}
\overrightarrow{a}\cdot\overrightarrow{b}
= \sum_{i=1}^{n}a_i b_i
= a_1b_1+a_2b_2+\cdots+a_nb_n
\label{eq:dot-product}
\end{equation}

\begin{equation}
P
= \frac{TP}{TP+FP}
= \frac{\sum_{i=1}^{N}\mathbf{1}\!\left(y_i=1\land\hat{y}_i=1\right)}
{\sum_{i=1}^{N}\mathbf{1}\!\left(\hat{y}_i=1\right)}
\label{eq:precision}
\end{equation}

\begin{equation}
R
= \frac{TP}{TP+FN}
= \frac{\sum_{i=1}^{N}\mathbf{1}\!\left(y_i=1\land\hat{y}_i=1\right)}
{\sum_{i=1}^{N}\mathbf{1}\!\left(y_i=1\right)}
\label{eq:recall}
\end{equation}

\begin{equation}
F_1
= 2\frac{PR}{P+R}
\label{eq:f1}
\end{equation}

\begin{equation}
J(A,B)
= \frac{|A\cap B|}{|A\cup B|}
\label{eq:jaccard}
\end{equation}

\begin{equation}
\mathrm{Accuracy}
= \frac{TP+TN}{TP+FP+FN+TN}
\label{eq:accuracy}
\end{equation}

\begin{equation}
\mathrm{Precision}
= \frac{TP}{TP+FP}
\label{eq:precision-simple}
\end{equation}

\begin{equation}
\mathrm{Recall}
= \frac{TP}{TP+FN}
\label{eq:recall-simple}
\end{equation}

\begin{equation}
F_1
= \frac{2(\mathrm{Precision})(\mathrm{Recall})}
{\mathrm{Precision}+\mathrm{Recall}}
\label{eq:f1-simple}
\end{equation}

\begin{equation}
\mathrm{TPR}
= \frac{\mathrm{True\ Positives}}
{\mathrm{True\ Positives}+\mathrm{False\ Negatives}}
\label{eq:tpr}
\end{equation}

\begin{equation}
\mathrm{FPR}
= \frac{\mathrm{False\ Positives}}
{\mathrm{False\ Positives}+\mathrm{True\ Negatives}}
\label{eq:fpr}
\end{equation}

\begin{equation}
\mathrm{AUC}
= \int_0^1 \mathrm{TPR}(\mathrm{FPR})\,d(\mathrm{FPR})
\label{eq:auc}
\end{equation}

\begin{equation}
\mathrm{TF\!-\!IDF}(t,d,D)
= \mathrm{TF}(t,d)\,\mathrm{IDF}(t,D)
\label{eq:tfidf}
\end{equation}

\end{document}